%% file: main.tex
\documentclass[10pt,twocolumn,letterpaper]{article}

\usepackage[pagenumbers]{cvpr} % To force page numbers, e.g. for an arXiv version

\input{preamble}
\setkeys{Gin}{draft=false}

\definecolor{cvprblue}{rgb}{0.21,0.49,0.74}
\usepackage[pagebackref,breaklinks,colorlinks,allcolors=cvprblue]{hyperref}

\def\paperID{*****} % *** Enter the Paper ID here
\def\confName{3DV\xspace}
\def\confYear{2027\xspace}

\title{Capturing Uncertainty in Human Motion
for Representation Learning in Soccer}

\author{
  Yizhou Xu\textsuperscript{1,2} \quad 
  Lars Bretzner\textsuperscript{2} \quad 
  Tiesheng Wang\textsuperscript{2} \quad 
  Atsuto Maki\textsuperscript{1} \\
  \textsuperscript{1}KTH Royal Institute of Technology \quad  
  \textsuperscript{2}EA Sports TRACAB 
  \thanks{
    This work was conducted as academic research at KTH Royal
    Institute of Technology in collaboration with Electronic Arts.
  }
  \\
  \textsuperscript{1}{\tt\small\{yizhoux,atsuto\}@kth.se} \quad
  \textsuperscript{2}{\tt\small\{lbretzner,tiewang\}@ea.com}
}

\begin{document}
\maketitle

\input{sec/0_abstract} 
\input{sec/1_introduction}

\input{sec/2_related_work}

\input{sec/3_method}

\input{sec/4_experiments}

\input{sec/5_conclusion}

\newpage
\input{sec/6_acknowledgement}

{
    \small
    \bibliography{reference, references}
    \bibliographystyle{ieeenat_fullname}
    
}

\input{sec/X_suppl}

\end{document}

%% file: preamble.tex
\DeclareUnicodeCharacter{21E4}{\ensuremath{\Leftarrow}}

\usepackage[utf8]{inputenc} % allow utf-8 input
\usepackage[T1]{fontenc}    % use 8-bit T1 fonts
\usepackage{url}            % simple URL typesetting
\usepackage{booktabs}       % professional-quality tables
\usepackage{amsfonts}       % blackboard math symbols
\usepackage{nicefrac}       % compact symbols for 1/2, etc.
\usepackage{microtype}      % microtypography
\usepackage{xcolor}         % colors

\usepackage{graphicx}
\usepackage{subcaption}
\usepackage{multirow}
\usepackage{colortbl}
\usepackage{bm}
\usepackage{amsmath}
\usepackage{amssymb}
\usepackage{mathtools}
\usepackage{amsthm}
\usepackage{algorithm}
\usepackage{algorithmic}
\usepackage{float}
\usepackage{lineno}

%% file: sec/0_abstract.tex
\begin{abstract}
This paper presents a self-supervised representation learning framework for understanding 3D skeleton-based human motion in soccer, using future motion prediction as the learning objective. 
Since human motion is inherently uncertain, accounting for multiple plausible futures is essential for capturing the underlying motion dynamics and learning effective representations.
To this end, we introduce a conditioning module for motion prediction that models a probabilistic distribution over discretized future motions in 3D Euclidean space, learning multimodality with explicit supervision from future trajectories.
Experiments on large-scale soccer player tracking data show that our approach substantially improves motion prediction accuracy.
Moreover, the learned representations effectively transfer to multiple soccer downstream applications, 
demonstrating strong cross-task generalization.
\end{abstract}

%% file: sec/1_introduction.tex
% \vspace{-10pt}
\section{Introduction}

\begin{figure}[t]
    % \vspace{-5pt} 
    \centering
    \includegraphics[width=\linewidth]{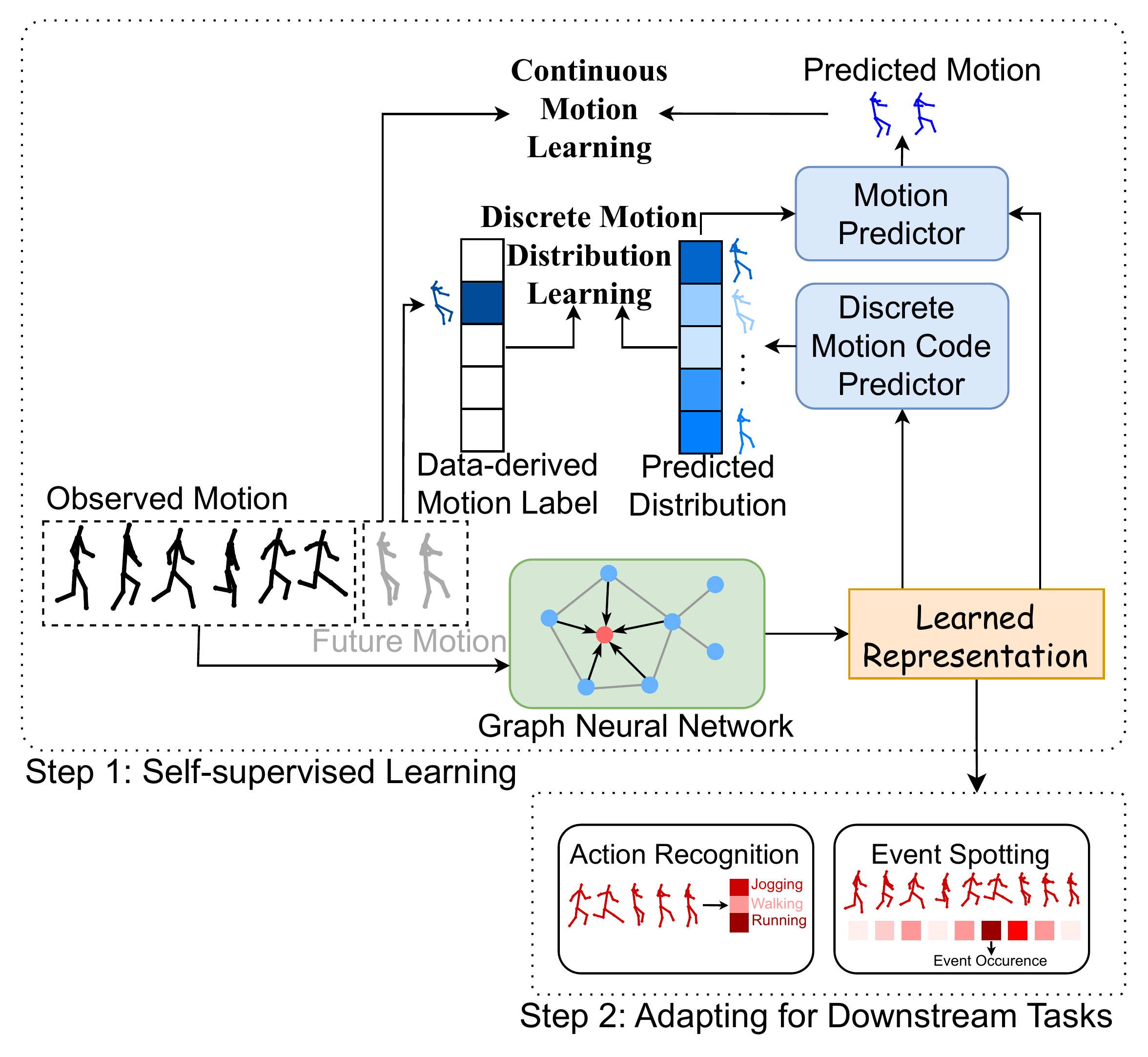}
    \caption{\textit{Overview of our self-supervised framework for human motion representation learning.}
    We introduce a conditioning module for motion prediction that learns a discrete distribution over future motions, enabling representations to capture uncertainty and adapt effectively to downstream tasks.}
    \label{fig:abstract}
    \vspace{-12pt}
\end{figure}

Human motion understanding is a fundamental problem in computer vision, where 3D skeleton data is widely used due to its robustness to environmental variations and lightweight representation. In recent years, skeletal representations have demonstrated strong performance in tasks such as motion prediction \cite{Mao2019LearningPrediction,Mao2020HistoryAttention,Ma2022ProgressivelyPrediction} and action recognition \cite{Yan2018,Shi2019Two-StreamRecognition,Chen2021Channel-WiseRecognition}, enabling applications in human-robot interaction~\cite{Koppula2016AnticipatingResponse,Gui2018TeachingMotion}, autonomous driving~\cite{Paden2016AVehicles}, and sports analytics~\cite{Ingwersen2023SportsPoseDataset,Yeung2024AutoSoccerPose:Movements}.

Despite their effectiveness, skeleton-based methods still face notable challenges in practical applications. 
Most existing approaches rely on lab-captured data, whereas real-world motions are often more diverse and less constrained. 
In addition, real-world systems often require a unified pipeline for multiple motion understanding tasks. Although recent work has begun to explore shared skeleton representations transferable across tasks~\cite{Yang2026PRISM:Representation}, most existing research remains task-specific due to variations in skeleton definitions across benchmarks for different tasks \cite{AMASS:ICCV:2019,Shahroudy2016NTUAnalysis, Wang2025HeterogeneousLearning}.
As a result, practitioners often have to maintain separate models for different tasks, complicating deployment.

To bridge these gaps, we study self-supervised representation learning from unlabeled 3D skeleton tracking data in soccer,
using motion prediction as the learning objective. 
Soccer offers a challenging yet well-suited setup for our purpose. Compared with lab environments, soccer player motions are faster, more diverse, and inherently multimodal. Soccer applications require multiple downstream tasks, such as action recognition and event detection. 
Moreover, a consistent skeleton definition can be used across tasks, making soccer a compelling domain for representation learning.

Motion prediction forecasts future pose sequences from historical observations, using future motion as supervision without requiring manual annotations. It has been widely studied with recent approaches typically based on graph convolutional networks (GCNs) \cite{Mao2019LearningPrediction,Dang2021MSR-GCN:Prediction,Ma2022ProgressivelyPrediction}. 
However, existing methods primarily prioritize prediction accuracy, leaving the potential of motion prediction as a self-supervised objective for representation learning underexplored.

Using motion prediction as a self-supervised objective introduces two key challenges.
Motion prediction is commonly formulated as a regression task, assuming a deterministic future given the observations. However, human motion is inherently stochastic and multimodal, 
making uncertainty modeling essential for learning representations that encode the underlying dynamics of human motion.
In addition, existing models typically process the entire sequence without frame-level supervision, which may limit their effectiveness for frame-level downstream tasks.

To address these challenges, we propose learning a discrete distribution over possible future motion modes, as illustrated in Figure~\ref{fig:abstract}. Specifically, we introduce a codebook that discretizes future motions at a fixed prediction horizon in 3D Euclidean space, mapping each ground-truth motion to a discrete code that provides explicit supervision for distribution learning. This supervision is crucial for stable and efficient distribution learning and prevents mode collapse in the codebook. By predicting probabilities rather than a single outcome, the model can capture multiple plausible futures and avoid producing an average motion.
Furthermore, we introduce frame-level supervision for motion prediction, requiring motion predictions at each frame rather than only at the end of a sequence. This design provides dense supervision throughout the sequence, which enables more efficient learning and is essential for learning frame-wise representations needed by frame-level downstream tasks.

We evaluate our approach on two soccer player motion datasets: the WorldPose dataset from the FIFA World Cup 2022 \cite{Jiang2024WorldPose:Estimation} and a proprietary dataset. 
Results show that learning a discrete future motion distribution significantly improves motion prediction performance. We further fine-tune the pre-trained model on two downstream tasks with different temporal granularities: sequence-level action recognition and frame-level shot spotting.
% , which requires temporal localization of shot events. 
Across both tasks, the learned representations yield consistent performance gains, indicating effective generalization to diverse applications.

In summary, our contributions are as follows:
% (1) We propose a self-supervised framework for learning human motion representations in soccer, using frame-level motion prediction as the learning objective;
% (2) We introduce discrete future distribution learning to model uncertainty in motion prediction, which enriches the learned motion representations;
% (3) We demonstrate on real-world soccer datasets that the proposed approach significantly improves motion prediction and generalizes to diverse downstream tasks.

\begin{itemize}
    \item We propose a self-supervised framework for learning human motion representations in soccer, using frame-level motion prediction as the learning objective;
    \item We introduce discrete future distribution learning to model uncertainty in motion prediction, which enriches the learned motion representations;
    \item We demonstrate on real-world soccer datasets that the proposed approach significantly improves motion prediction and generalizes to diverse downstream tasks.
\end{itemize}

%% file: sec/2_related_work.tex
\vspace{-5pt}
\section{Related Work}

\subsection{Human Motion Prediction}
% \vspace{-5pt}

Human motion prediction aims to forecast future motion from observations. 
Early approaches modeled temporal dynamics with recurrent neural networks (RNNs) \cite{Fragkiadaki2015RecurrentDynamics,Jain2016Structural-RNN:Graphs,Martinez2017OnNetworks,Chiu2019Action-agnosticForecasting,Gopalakrishnan2019APrediction} or convolutional neural networks (CNNs)~\cite{Butepage2017DeepClassification,Li2018ConvolutionalDynamics}.
More recent methods represent motion as a dynamic graph and employ GCNs~\cite{Mao2019LearningPrediction,Li2020DynamicPrediction,Mao2020HistoryAttention,Dang2021MSR-GCN:Prediction,Sofianos2021Space-Time-SeparableForecasting,Ma2022ProgressivelyPrediction,Li2022SymbioticPrediction,Li2022Skeleton-PartedPrediction,Wang2024GCNext:Prediction} to capture structured inter-joint relationships. 
Multilayer perceptron (MLP)-based approaches have also been explored, including pure feedforward architectures \cite{Guo2023BackPrediction} and methods incorporating physics-inspired priors \cite{Xu2023EqMotion:Reasoning,Wang2024GCNext:Prediction,Wei2026ProgressiveModels}.
Despite their success, most existing methods treat motion prediction as deterministic regression, emphasizing mean prediction accuracy rather than modeling distributions over possible futures.

Another line of work focuses on diverse human motion prediction, aiming to generate multiple plausible futures from a single observation. Most approaches address this using generative models, including generative adversarial networks (GANs) \cite{Barsoum2018HP-GAN:GAN,Kundu2019BiHMP-GAN:GAN}, variational autoencoders (VAEs) \cite{Yan2018MT-VAE:Dynamics,Cai2021AAuto-Encoder}, and diffusion- and flow-based models \cite{Saadatnejad2022AWild,Chen2023HumanMAC:Prediction,Barquero2023BeLFusion:Prediction,Sun2024CoMusion:Diffusion,Curreli2025NonisotropicPrediction,Ma2026Gaussian-MixturePrediction}. In these methods, diversity is typically achieved by sampling from a continuous latent distribution, often Gaussian. While various sampling strategies have been proposed to improve diversity and prediction quality~\cite{Aliakbarian2020APrediction,Yuan2020DLow:Prediction,Mao2021GeneratingPrediction,Dang2022DiverseSpace,Xu2022DiverseAnchors,Kim2024MotionNetworks,Hosseininejad2025MotionMap:Forecasting}, these methods primarily focus on generating diverse samples rather than learning a conditional probability distribution over future modes, which is essential for capturing multimodal motion dynamics.
Moreover, diffusion- and flow-based methods are typically optimized for generation and lack a standardized latent representation that can be readily used for downstream tasks.
In contrast, our approach explicitly learns a reusable representation and models the distribution of future motion, enabling learned representations to account for multimodality.
% In contrast, our approach explicitly models the distribution of future motion, enabling learned representations to account for multimodality.

% \vspace{-5pt}
\subsection{Skeleton-based Action Recognition}
% \vspace{-5pt}
Skeleton-based action recognition is a well-studied problem in human motion understanding~\cite{Wang2014Cross-viewRecognition, Shahroudy2016NTUAnalysis, Yan2018, Liu2019NTUUnderstanding, Shi2019Two-StreamRecognition}, with supervised GCN-based models achieving strong performance \cite{Liu2020DisentanglingRecognition,Chen2021Channel-WiseRecognition,Chi2022InfoGCN:Recognition,Lee2023HierarchicallyRecognition,Zhou2024BlockGCN:Recognition,Liu2025RevealingRecognition}. Although some works explore unsupervised learning to reduce dependence on labeled data \cite{Mao2022CMD:Distillation,Guo2022ContrastiveRecognition,Zhang2022ContrastiveLearning,Kim2022Global-LocalLearning,Mao2023MaskedLearners,Franco2023HYperbolicRepresentations,Wu2024MacDiff:Diffusion,Abdelfattah2024S-JEPA:Recognition,Sun2026ExploringRecognition}, they typically rely on datasets designed for classification and treat them as unlabeled data during training, which limits data scale and motion diversity.
In contrast, large-scale unlabeled data is readily accessible in soccer through player tracking, while annotations are scarce. We therefore adopt motion prediction for self-supervised learning and adapt the learned representations for action recognition as a downstream task.

% \vspace{-5pt}
\subsection{Discrete Latent Modeling of Skeleton Motion}
% \vspace{-5pt}

Discrete latent representations for skeleton motion have been explored mainly in the context of text-to-motion generation \cite{Guo2022TM2T:Texts,Zhong2023AttT2M:Mechanism,Zhang2023GeneratingRepresentations,Jiang2023MotionGPT:Language,Kong2023Priority-CentricSpace,Chi2024M2D2M:Models,Guo2024MoMask:Motions,Chen2025TheMotion}. These works typically learn a discrete codebook in the latent space using reconstruction-based objectives such as VQ-VAE \cite{vandenOord2017NeuralLearning}, where discretization serves to bridge continuous motion with discrete language tokens. While effective for motion generation, such codebooks are learned without explicit supervision, making them weakly constrained and prone to collapse or under-utilization.
% and are therefore weakly constrained during training, which can lead to codebook collapse or under-utilization.
A small number of works sample from a discrete latent space for motion prediction~\cite{Salzmann2022Motron:Forecasting,Xu2024LearningPrediction}, but in the absence of explicit supervision, these methods suffer from similar limitations.
In contrast, our approach learns a discrete distribution in 3D motion space, using future motion as explicit supervision, resulting in more stable and reliable representations.

%% file: sec/3_method.tex
% \vspace{-5pt}
\section{Method}
\label{sec:method}
% \vspace{-5pt}

In this section, we present our self-supervised learning framework for skeleton-based motion understanding.
We first formulate the problem in Section~\ref{sec:problem-formulation}, then describe the graph-based modeling backbone in Section~\ref{sec:spatio-temporal-graph}, the proposed discrete distribution learning (DDL) framework in Section~\ref{sec:discrete-module}, and its adaptation to downstream tasks in Section~\ref{sec:gtn-on-downstream}.

% \vspace{-5pt}
\subsection{Problem Formulation}
\label{sec:problem-formulation}
% \vspace{-5pt}

We learn human motion representations using frame-level motion prediction as a self-supervised objective. Given an observed motion sequence of $N$ frames, $\bm{X}_{1:N} = \{\bm{x}_t\}_{t=1}^{N}$, where $\bm{x}_t \in \mathbb{R}^{J \times 3}$ denotes a human pose represented by the 3D coordinates of $J$ joints, the model predicts future motion at each observed frame $t$. Specifically, for each $t \in \{1,\dots,N\}$, the model predicts the next $T$ future frames $\hat{\bm{X}}_{t+1:t+T}$ from the observed prefix $\bm{X}_{1:t}$, supervised by the ground-truth sequence $\bm{X}_{t+1:t+T}$.

% \vspace{-5pt}
\subsection{Graph Modeling of Skeletal Motion}
\label{sec:spatio-temporal-graph}
% \vspace{-5pt}
\subsubsection{Spatio-temporal Graph Construction}
% \vspace{-5pt}

Given a skeletal motion sequence $\mathbf{X}_{1:N}$, we represent it as a spatio-temporal graph with the joints as nodes and edges capturing the natural human body structure and its temporal dynamics.
Specifically, each node corresponds to a joint at a specific frame. 
To support frame-level tasks, we introduce an abstract player node at each frame, which is connected to all joints in that frame to aggregate joint-level information into a frame-level representation.

Edges are constructed to capture the skeletal structure and temporal evolution, 
with directed edges enforcing temporal causality to support frame-level learning.
Specifically, the edges include 
    (a) semantic skeletal connections (bones) within a frame; 
    (b) temporal connections linking the same joints across consecutive frames by directed edges;
    (c) player node connections, where all joints in a frame are connected to the player node.
All temporal connections are directed from past to future to prevent future information leakage. In particular, each player node is temporally connected to player nodes at all future frames.
With this design, any two nodes in the graph are separated by at most three hops, enabling efficient information flow across space and time.

\noindent\textbf{Node and Edge Features.}\quad
Each node feature encodes the 3D coordinates of the joint together with a learnable joint-type embedding. Each edge is similarly associated with a feature that encodes its connection type (e.g., skeletal or temporal connection) and the geometric distance between the connected nodes. These features provide the input representations for graph-based message passing. 
Details of the node and edge features are provided in Appendix~\ref{appendix:graph-construction}.

% \vspace{-8pt}
\subsubsection{Graph Transformer Network}
% \vspace{-5pt}

To capture spatio-temporal dependencies in the constructed graph, we use a Graph Transformer Network (GTN) as a graph neural network (GNN) backbone for representation learning. 
% GTN adopts modern transformer designs for message passing as a GNN.
GTN adopts modern transformer components into graph message passing.
The causal graph structure ensures that the node representations produced by GTN encode only past information, making them well-suited for frame-level motion prediction. 
Implementation details of GTN are provided in Appendix~\ref{appendix:gtn}.

\begin{figure*}[t]
    \centering
    % \vspace{-10pt}
    \includegraphics[width=0.85\linewidth]{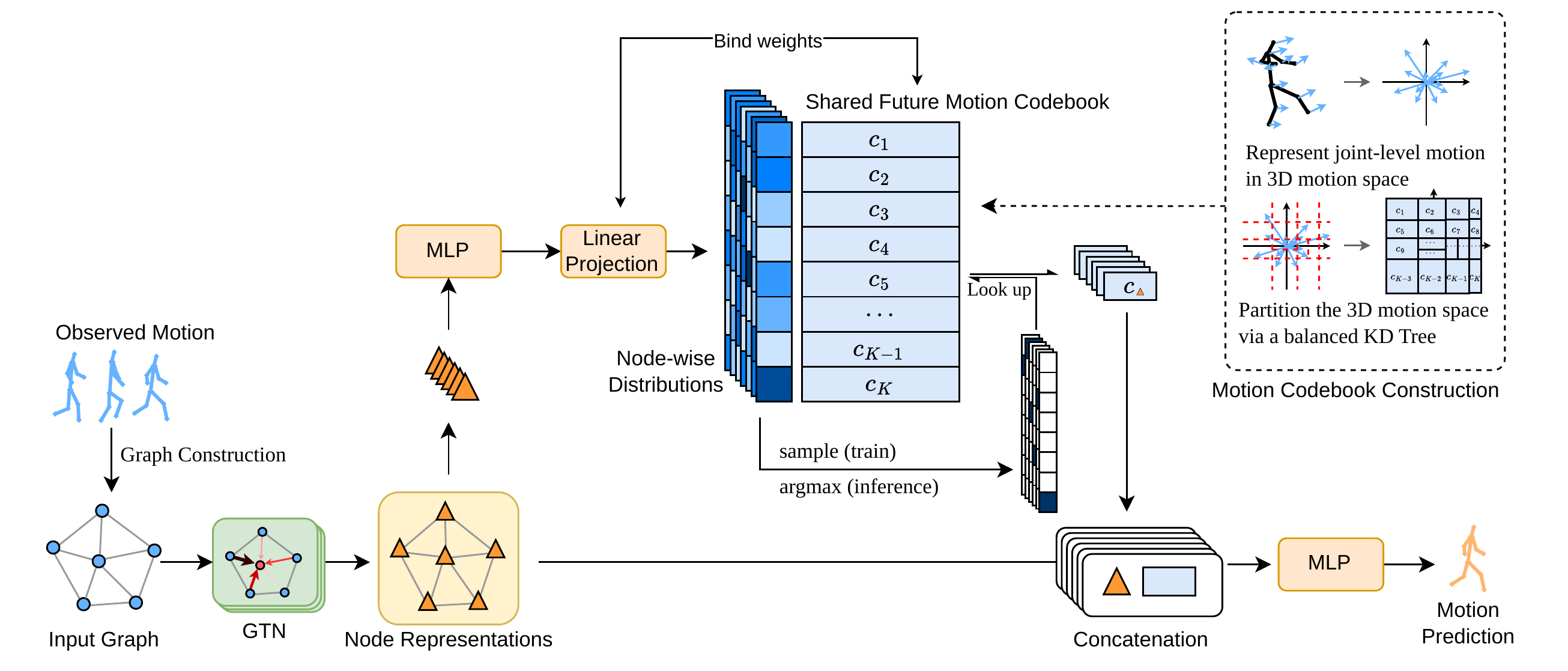}
    \caption{\textit{Overview of our discrete distribution learning approach.} 
    For each node representation produced by the Graph Transformer Network (GTN), the model predicts a categorical distribution over a shared motion codebook, which is constructed by discretizing the joint-level 3D motion space. A code is then sampled from this distribution (argmax during inference), and its embedding is concatenated with the node representation for final motion prediction.}
    \label{fig:codebook-overview}
    \vspace{-10pt}
\end{figure*}

\noindent\textbf{GTN for Motion Prediction.}\quad
GTN outputs a node representation for each joint per frame, which is used to predict future motions of this joint over the next $T$ frames via a shared linear projection. The model is trained with a motion prediction loss $\mathcal{L}_{\mathrm{motion}}$, defined as the mean squared error (MSE) between the predicted and ground-truth motion. See Appendix~\ref{appendix:gtn-mse-loss} for the formal formulation.

% \vspace{-5pt}
\subsection{DDL for Future Motion}
\label{sec:discrete-module}
% \vspace{-5pt}

To model uncertainty in motion prediction, we introduce a future motion codebook that discretizes the 3D future motion space. This enables the model to learn a discrete distribution over future motion codes, which subsequently conditions the final motion prediction. We describe the codebook construction in Section~\ref{sec:codebook-construction}, and the discrete distribution learning (DDL) mechanism in Section~\ref{sec:codebook-framework}.

% \vspace{-5pt}
\subsubsection{Codebook Construction}
\label{sec:codebook-construction}
% \vspace{-5pt}

We introduce a future motion codebook to enable distribution learning over possible future motions.
The codebook defines a deterministic mapping from continuous motion to a finite set of discrete codes, allowing the model to predict a categorical distribution over the codes. 
Specifically, we discretize the joint-level 3D motion space at a fixed future offset of $t_f$ frames and keep the codebook unchanged thereafter to ensure stable learning. 
During training, ground-truth future motions are mapped to their corresponding codes to serve as supervision signals, enabling self-supervised learning of the future motion distribution.
Since discretization is performed at the joint level with each code representing a 3D joint displacement, a skeleton motion with $J$ joints is encoded by $J$ motion codes. 
Given a codebook of size $K$, this yields a combinatorial space of size $K^J$ for each skeleton motion, enabling expressive modeling of skeleton dynamics.

To ensure balanced code utilization, we employ a balanced k-dimensional tree (KD-tree) to partition the 3D motion space into $K$ regions. As shown in Figure~\ref{fig:codebook-overview} on the right side, we first collect motion samples from the training set to represent the 3D motion space,  
where each sample corresponds to a 3D displacement of a joint at the $t_f$-th frame. 
Starting from all samples as a single region, the algorithm recursively selects the most populated region, splits it along the dimension with the largest variance at the median, and repeats this process until $K$ regions are obtained.
This procedure produces a set of regions with comparable sample counts, where each region defines a motion code.
Full algorithmic details are provided in Appendix~\ref{appendix:kd-tree-construction}.

By splitting along directions with the highest variation while enforcing balanced partitions, the KD-tree captures the structure of the motion distribution and yields motion codes with near-uniform usage frequency, which is crucial for effective distribution learning. The resulting codebook provides an efficient mapping from continuous motions to discrete codes via tree traversal, enabling stable supervision with minimal additional costs during training.

\input{sec/table_1}

% \vspace{-6pt}
\subsubsection{DDL for Motion Prediction}
\label{sec:codebook-framework}
% \vspace{-2pt}

DDL captures future motion uncertainty by predicting a categorical distribution over the motion codebook.
The final motion predictor is then conditioned on a code sampled from this distribution, enabling uncertainty-aware motion predictions.
As shown in Figure~\ref{fig:codebook-overview}, given GTN-produced node representations, the model infers a distribution over motion codes and samples a code accordingly. The corresponding code embedding is then fused with the node representation for final motion prediction.
Note that using the code embedding alone for motion prediction would discard fine-grained continuous motion information due to discretization, which can be detrimental for accurate motion prediction. Therefore, we use the code embedding as a conditioning signal rather than a standalone representation.
To prevent code embedding collapse, we tie the code embedding matrix to the projection layer used for code classification. This weight sharing enforces embedding separation, since identical embeddings would prevent correct classification of motion codes.

Formally, for a node $u$, let the node representation produced by the $L$-layer GTN be
$\bm{h}_u^{(L)} \in \mathbb{R}^{d}$.
The code embedding matrix is
$\mathbf{E} = \{\mathbf{e}_{c_k}\}_{k=1}^K \in \mathbb{R}^{K \times d}$,
where each $\bm{e}_{c_k} \in \mathbb{R}^{d}$ is the embedding of code $c_k$.
The probability distribution over the codebook is predicted by
\begin{equation}
\label{eqn:dist}
\bm{p}_{u} = \mathrm{softmax}\big(\mathrm{proj}\big(\mathrm{MLP}_{\mathrm{code}}(\bm{h}_u^{(L)})\big) / \tau\big),
\end{equation}
where $\mathrm{proj}(\cdot)$ is a linear projection from $\mathbb{R}^{d}$ to $\mathbb{R}^{K}$, with its weights tied to the code embedding matrix $\mathbf{E}$. $\mathrm{MLP}_{\mathrm{code}}$ is a two-layer MLP mapping from $\mathbb{R}^d$ to $\mathbb{R}^d$, with GELU activation and hidden dimension $d$. $\tau$ is a temperature controlling the distribution sharpness. We use $\tau = 1$ for training.

A motion code $c_u$ is then sampled from $\bm{p}_u$ during training,
whereas during inference we select the most probable code.
% Let $\bm{e}_{c_u} \in \mathbb{R}^{d}$ denote the selected motion code embedding.
The selected code embedding $\bm{e}_{c_u} \in \mathbb{R}^{d}$ is subsequently fused with the node representation $\bm{h}_u^{(L)}$ via concatenation followed by an MLP:
\begin{equation}
\bm{h}_u^{\mathrm{fused}}
=
\mathrm{MLP}_{\mathrm{fuse}}\!\left(
[\bm{h}_u^{(L)} \,\Vert\, \bm{e}_{c_u}]
\right),
\end{equation}
where $\mathrm{MLP}_{\mathrm{fuse}}$ is a two-layer MLP from $\mathbb{R}^d$ to $\mathbb{R}^d$ with GELU activation and hidden dimension $d$.

A learnable projection $\mathbf{W}_p: \mathbb{R}^d \rightarrow \mathbb{R}^{T \times 3}$ is then used to predict joint-level motion for node $u$ from the fused representation $\bm{h}_u^{\mathrm{fused}}$:
\begin{equation}
    \hat{\bm{m}}_{u,1:T}
    =
    \mathbf{W}_p \bm{h}_{u}^{\mathrm{fused}}
    \in \mathbb{R}^{T \times 3}.
\end{equation}

\noindent\textbf{Training Objective.}\quad
Given the ground-truth future motion at the $t_f$-th frame, we obtain the corresponding motion code via KD-tree traversal and use it as supervision. Code prediction is treated as a classification task and optimized using a standard cross-entropy loss $\mathcal{L}_{\mathrm{code}}$.
The overall training objective combines the motion loss $\mathcal{L}_{\mathrm{motion}}$ and the code loss $\mathcal{L}_{\mathrm{code}}$ with a coefficient $\alpha$ controlling the contribution of the DDL objective:
\begin{equation}
\label{eqn:loss}
\mathcal{L}
=
\mathcal{L}_{\mathrm{motion}}
+
\alpha\, \mathcal{L}_{\mathrm{code}}.
\end{equation}

% \vspace{-14pt}
\subsection{Adaptation to Downstream Tasks}
\label{sec:gtn-on-downstream}
% \vspace{-5pt}

Our model is straightforwardly adaptable to downstream tasks. Given the learned representations, we perform task-specific adaptation by attaching a lightweight prediction head and fine-tuning the network using only the GTN backbone, without the DDL module. DDL is used solely during pre-training to enhance representation quality, and is not required for adaptation. We demonstrate this transferability on two downstream tasks: action recognition and shot spotting.

\noindent\textbf{Action Recognition.}\quad
Action recognition is a sequence-level classification task. We fine-tune the model by adding a linear classification head on top of the graph representation, obtained by mean-pooling the node representations from the final GTN layer across all nodes in the graph. All parameters are optimized using standard cross-entropy loss.

\noindent\textbf{Shot Spotting.}\quad
Shot spotting aims to temporally localize the frame at which a shot occurs within a motion sequence, or determine that no shot is present. We formulate it as a frame-level prediction task, where the model predicts a confidence score for each frame, indicating the likelihood of a shot.
We fine-tune the model by applying a linear layer with sigmoid activation to the player node representation at each frame, producing a confidence score in $[0,1]$. The model is trained with binary cross-entropy loss.
During inference, the model outputs frame-wise confidence scores and identifies the shooting frame as the temporal peak, applying a threshold to confirm shot presence.

%% file: sec/table_1.tex
\begin{table*}[t]
% \vspace{-10pt}
\centering
\caption{
\textit{Motion prediction results.} 
We report MPJPE from 80\,ms to 1000\,ms;
lower is better.
}
\label{tab:motion-prediction-merged}
\resizebox{\textwidth}{!}{%
\begin{tabular}{@{}l|llllllll||llllllll@{}}
\toprule
\multirow{2}{*}{Model}
& \multicolumn{8}{c||}{MPJPE (mm) $\downarrow$ on WorldPose}
& \multicolumn{8}{c}{MPJPE (mm) $\downarrow$ on ProSoccer} \\
\cmidrule(lr){2-9} \cmidrule(lr){10-17}
& 80ms & 160ms & 320ms & 400ms & 560ms & 720ms & 880ms & 1000ms
& 80ms & 160ms & 320ms & 400ms & 560ms & 720ms & 880ms & 1000ms \\
\midrule
Zero %(Repeating last frame)
& 200.4 & 397.3 & 784.6 & 976.5 & 1356.5 & 1734.4 & 2109.4 & 2387.2
& 148.1 & 297.4 & 587.9 & 731.1 & 1015.0 & 1296.9 & 1576.5 & 1783.5 \\
\midrule
LTD \cite{Mao2019LearningPrediction}
& 19.0 & 39.3 & 72.5 & 89.2 & 142.2 & 217.6 & 328.3 & 436.0
& 15.7 & 28.7 & 54.7 & 68.0 & 102.6 & 148.4 & 209.4 & 267.1 \\
HisRep \cite{Mao2020HistoryAttention}
& 18.7 & 39.1 & 72.0 & 88.6 & 132.9 & 195.3 & 287.4 & 376.3
& 15.3 & 28.3 & 54.1 & 67.4 & 100.2 & 143.7 & 202.2 & 257.4 \\
MSR-GCN \cite{Dang2021MSR-GCN:Prediction}
& 23.7 & 46.5 & 81.3 & 100.3 & 161.2 & 238.5 & 351.5 & 459.4
& 24.9 & 43.6 & 76.6 & 94.6 & 144.2 & 205.0 & 289.9 & 362.9 \\
PGBIG \cite{Ma2022ProgressivelyPrediction}
& 17.3 & 37.8 & 69.0 & 85.6 & 135.1 & 202.7 & 305.6 & 405.8
& 17.8 & 34.1 & 64.7 & 80.7 & 121.6 & 170.7 & 238.2 & 300.9 \\
SiMLPe \cite{Guo2023BackPrediction}
& 19.7 & 43.7 & 83.5 & 103.5 & 158.2 & 232.7 & 341.2 & 445.5
& 17.7 & 34.3 & 66.6 & 83.4 & 124.7 & 176.7 & 245.0 & 308.7 \\
GCNext \cite{Wang2024GCNext:Prediction}
& 18.5 & 41.3 & 77.5 & 95.4 & 142.7 & 211.1 & 308.2 & 403.2
& 18.8 & 37.0 & 71.2 & 88.6 & 130.8 & 184.3 & 253.8 & 317.8 \\
\midrule
GTN (ours)
& 16.3 & 34.1 & 62.9 & 77.8 & 118.6 & 176.7 & 261.1 & 344.7
& 15.2 & 27.8 & 53.2 & 66.4 & 99.0 & 141.8 & 199.6 & 254.7 \\
GTN+DDL (ours)
& \textbf{15.6} & \textbf{32.5} & \textbf{61.0} & \textbf{76.2}
& \textbf{115.9} & \textbf{173.4} & \textbf{256.7} & \textbf{338.6}
& \textbf{13.8} & \textbf{24.4} & \textbf{49.2} & \textbf{62.3}
& \textbf{94.5} & \textbf{136.6} & \textbf{193.5} & \textbf{247.7} \\
\bottomrule
\end{tabular}
}
% \vspace{-10pt}
\end{table*}

%% file: sec/4_experiments.tex
% \vspace{-5pt}
\section{Experiments}
% \vspace{-5pt}

\subsection{Motion Prediction}
% \vspace{-5pt}
    \textbf{Data.}\quad
        % We use two datasets for motion prediction to validate our proposed method for representation learning.
        We evaluate motion prediction performance on two soccer player tracking datasets, with all data interpolated to 25 Hz. The observation window $N$ is 50 frames, and the prediction horizon $T$ is 10 frames.
        \textit{WorldPose}
        % \footnote{We use WorldPose under ETH Zurich's custom license: \url{https://worldpose.ait.ethz.ch/}.}
        ~\cite{Jiang2024WorldPose:Estimation} contains approximately 11.4 hours of player tracking data from 8 matches of the 2022 FIFA World Cup. Although introduced for pose estimation, we repurpose its ground-truth annotations for motion prediction. We hold out the \texttt{BRA-KOR} match for evaluation and use the remaining matches for training, resulting in 9.5 hours of training data and 1.9 hours of test data. 
        % Motions are provided in SMPL format and converted to our 21-joint skeleton definition.
        \textit{ProSoccer} is a proprietary dataset composed of commercial soccer player tracking data, with a substantially larger scale than WorldPose.

    \noindent\textbf{Experimental Details.}\quad
        % We use dataset-specific configurations and training setups. 
        We construct the future motion codebook by discretizing the motion at the 5th ($t_f = 5$)  prediction step. 
        We use dataset-specific configurations and training setups, with full details provided in Appendix~\ref{appendix:exp_details_motion_prediction}.

    \noindent\textbf{Metric.}\quad
        We evaluate motion prediction performance using Mean Per Joint Position Error (MPJPE), computed as the average Euclidean distance between predicted and ground-truth joint positions (lower is better). Results are reported at prediction horizons from 80\,ms to 1000\,ms, with longer horizons predicted autoregressively by the model.

\begin{figure*}[t]
    % \vspace{-10pt}
    \centering
    \begin{subfigure}[t]{0.48\textwidth}
        \centering
        \includegraphics[width=0.85\linewidth]{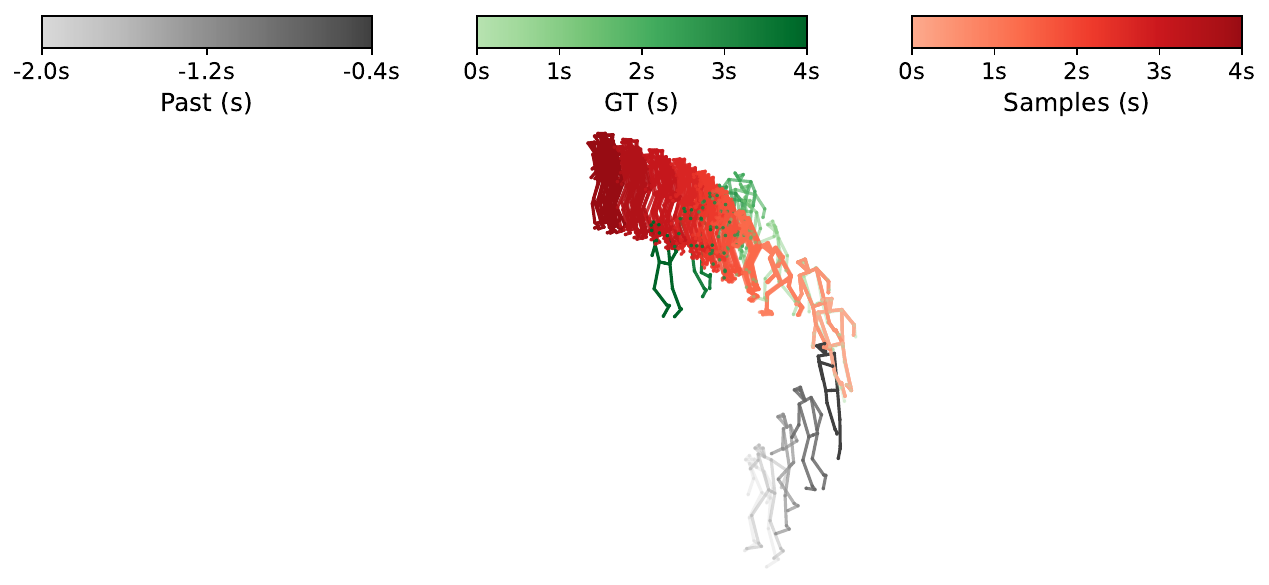}
        \caption{$\tau = 1.0$}
        \label{fig:motion_vis1a}
    \end{subfigure}
    \hfill
    \begin{subfigure}[t]{0.48\textwidth}
        \centering
        \includegraphics[width=0.85\linewidth]{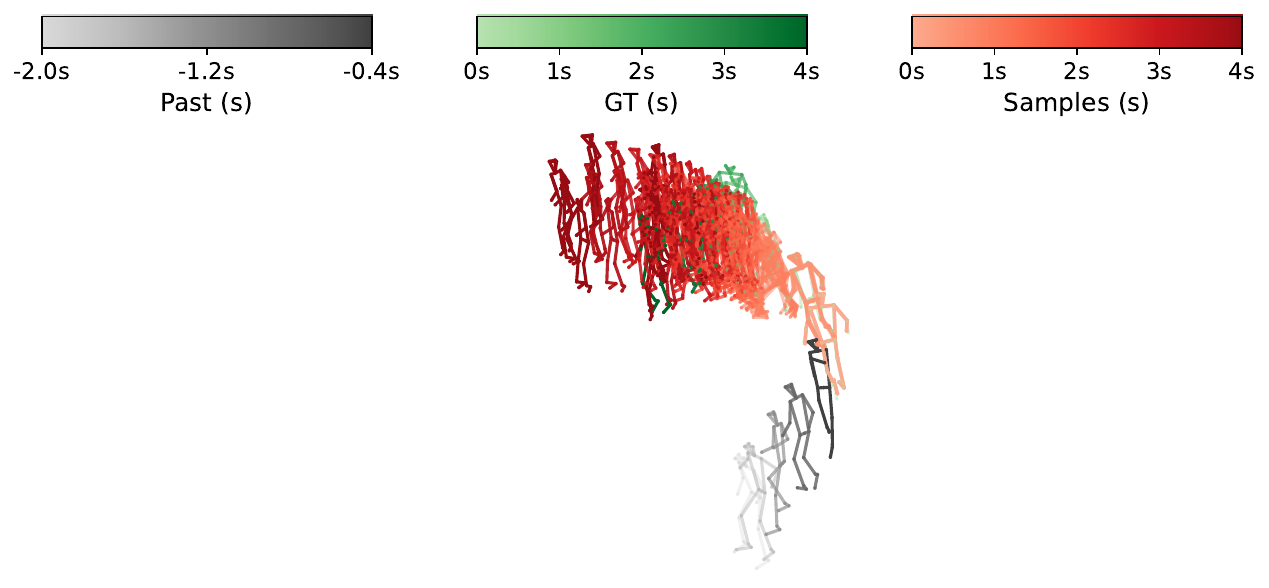}
        \caption{$\tau = 10.0$}
        \label{fig:motion_vis1b}
    \end{subfigure}
    \caption{
    \textit{Stochastic motion predictions.} 
    % 10 samples are predicted (red) at temperature $\tau = 1$ (a) and $\tau = 10$ (b). Grey: input; Green: ground truth; Color gradients indicate temporal progression.
    Ten motion samples are predicted at temperatures (a) $\tau=1$ and (b) $\tau=10$. Gray denotes the input sequence, green the ground truth, and red the predictions; color gradients indicate temporal progression.
    }
    \label{fig:motion_prediction_visualization1}
    \vspace{-5pt}
\end{figure*}

    \noindent\textbf{Baselines.}\quad
        We compare our method with six representative prior motion prediction approaches: 
        LTD~\cite{Mao2019LearningPrediction}, HisRep~\cite{Mao2020HistoryAttention}, 
        MSR-GCN~\cite{Dang2021MSR-GCN:Prediction}, PGBIG~\cite{Ma2022ProgressivelyPrediction}, 
        SiMLPe~\cite{Guo2023BackPrediction}, and GCNext~\cite{Wang2024GCNext:Prediction}. 
        We additionally include a naive baseline \textbf{Zero} that repeats the last observed frame.
        Details of motion prediction baselines are provided in Appendix~\ref{appendix:motion-prediction-baseline}.

    \noindent\textbf{Motion Prediction Results.}\quad
        Table~\ref{tab:motion-prediction-merged} reports motion prediction results on WorldPose and ProSoccer. 
        As shown in the table, the GTN backbone alone outperforms all baseline methods on both datasets, demonstrating its effectiveness for modeling soccer player motion. 
        Incorporating discrete distribution learning (DDL) further improves performance over the GTN backbone by a clear margin for both short-term and long-term predictions, highlighting the benefit of explicitly modeling future motion uncertainty.
        Although the codebook is constructed using motion at the 5th future frame (200\,ms at 25\,Hz), the performance improvement extends beyond the direct prediction window of 10 frames (400\,ms) under autoregressive evaluation. This indicates that DDL improves the overall quality of the learned motion representation, rather than biasing the model toward a specific temporal horizon.

    \noindent\textbf{Qualitative Diverse Motion Prediction Results.}\quad
        To examine the learned distribution, we sample joint motion codes from the learned distribution $\bm{p}_{u}$ (Equation~\ref{eqn:dist}) to enable stochastic motion prediction at inference. 
        Given a 2-second input (50 frames), we generate 10 future motions of 4 seconds (100 frames) and visualize them in Figure~\ref{fig:motion_prediction_visualization1} with two different temperatures.
        At $\tau = 1$ (Figure~\ref{fig:motion_vis1a}), the predictions exhibit some diversity, but remain concentrated around dominant modes. Increasing $\tau$ to 10 (Figure~\ref{fig:motion_vis1b}) produces more diverse samples, some of which are closer to the ground truth. This indicates that the learned codebook captures multimodal structure and uncertainty, though the prediction-only training objective leads to a sharp distribution. Simply increasing the temperature can reveal the learned latent multimodality without retraining. We provide a further quantitative evaluation showing this effect in Appendix~\ref{appendix:diverse_results_with_taus}. Nevertheless, our goal is representation learning rather than diverse motion generation. This experiment serves only as a qualitative demonstration of learned uncertainty.

\begin{figure}
    % \vspace{-10pt}
    \centering
    \includegraphics[width=\linewidth]{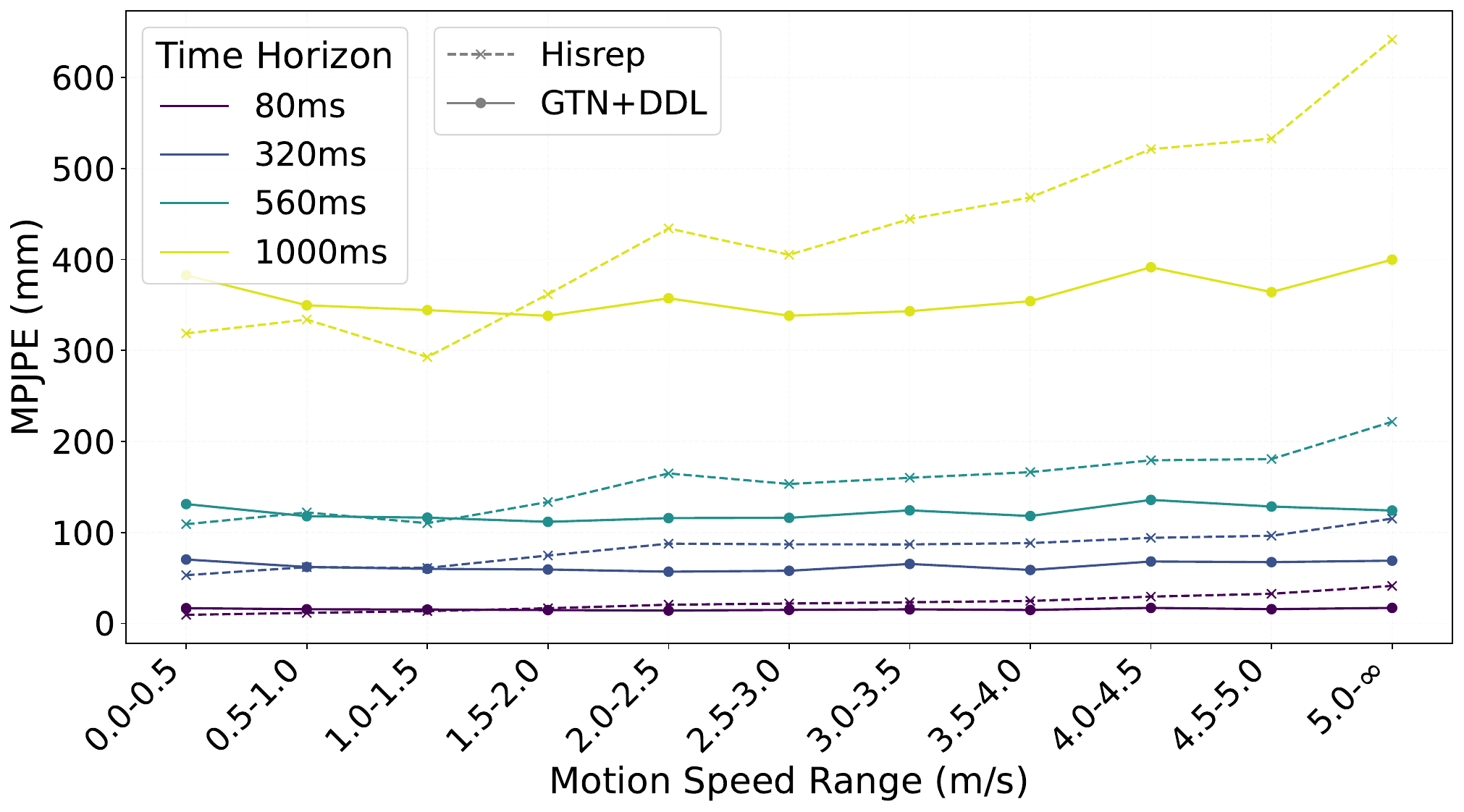}
    \caption{
    \textit{Motion prediction performance across different motion speeds on WorldPose.} We compare the GTN trained with DDL (solid) against HisRep (dashed).}
    \label{fig:range_eval_worldpose}
    \vspace{-12pt}
\end{figure}

    \noindent\textbf{Motion Prediction across Different Speeds.}\quad
        To evaluate robustness across different speeds, we group the test set by motion speed using 0.5\,m/s intervals, ranging from 0-0.5\,m/s to above 5\,m/s. Figure~\ref{fig:range_eval_worldpose} compares our GTN trained with DDL against the strongest baseline HisRep on WorldPose, reporting MPJPE for short-term (80\,ms, 320\,ms) and long-term (560\,ms, 1000\,ms) horizons.
        As shown in the figure, HisRep exhibits lower error at the lowest speed range, but its error increases rapidly as the motion speed grows. In contrast, our GTN+DDL exhibits more robust performance as speed increases, with significantly slower error growth. As a result, our approach outperforms HisRep by an increasing margin in high-speed scenarios, demonstrating strong robustness to fast motions. We observe a similar trend on ProSoccer in Appendix~\ref{appendix:range-eval-further} and further show that GTN plays an important role in the robustness to high-speed motion, with DDL providing consistent gains.

% \vspace{-5pt}
\subsection{Downstream Tasks}
% \vspace{-5pt}

    \subsubsection{Action Recognition}
    \label{sec:exp_action_recognition}
    % \vspace{-5pt}
 
        \textbf{Data.}\quad
            We evaluate the learned motion representations on two action recognition benchmarks. 
            \textit{WorldPoseAR}
            % \footnote{This benchmark is provided in the supplementary materials, and will be made public if this paper is accepted.} 
            is an 8-class soccer action recognition benchmark we constructed from WorldPose~\cite{Jiang2024WorldPose:Estimation}, containing seven off-ball movement actions and one ball-kicking action. \textit{SoccerAR} is an in-house benchmark with 35 action classes, covering both on-ball and off-ball actions. In both datasets, each sample is a 50-frame skeleton sequence with a single label, and the validation and test sets are class-balanced. Details of the action recognition benchmarks are provided in Appendix~\ref{sec:appendix_ar_benchmarks}.

        \noindent\textbf{Experimental Details.}\quad
            We fine-tune the motion prediction pre-trained model for action recognition, with full details provided in Appendix~\ref{appendix:ar_details}.
            We report test-set accuracy as the mean and standard deviation over three random seeds.

        \noindent\textbf{Baselines.}\quad
            We compare our approach with three representative skeleton-based action recognition methods: ST-GCN~\cite{Yan2018}, CTR-GCN~\cite{Chen2021Channel-WiseRecognition}, and BlockGCN~\cite{Zhou2024BlockGCN:Recognition}. 
            We additionally evaluate a GTN baseline trained from scratch using the same architecture as our pre-trained models. To compare with self-supervised learning approaches, we include MAMP~\cite{Mao2023MaskedLearners}.
            Details are provided in Appendix~\ref{appendix:downstream-baseline}.

        \noindent\textbf{Results.}\quad
            Table \ref{tab:action-recognition} reports action recognition results on WorldPoseAR and SoccerAR. The GTN initialized with GTN+DDL pre-trained weights achieves the best performance consistently with 96.21\% accuracy on WorldPoseAR and 64.15\% accuracy on SoccerAR, substantially outperforming all baselines, including the self-supervised learning method MAMP~\cite{Mao2023MaskedLearners}.
            % Compared with the self-supervised learning baseline MAMP~\cite{Mao2023MaskedLearners}, GTN+DDL improves accuracy by \(3.04\) percentage points on WorldPoseAR and \(5.03\) percentage points on SoccerAR.
            The same GTN trained from scratch shows substantially lower performance, indicating the importance of representation learning via motion prediction.
            % The model initialized from a plain pre-trained GTN yields strong but lower performance, consistent with its lower motion prediction accuracy in Table~\ref{tab:motion-prediction-merged} , demonstrating the benefit of DDL pre-trained representations for downstream tasks. 
            Pre-training with the plain GTN also performs strongly, but remains below the DDL pre-trained model. This is consistent with its lower motion prediction accuracy in Table~\ref{tab:motion-prediction-merged} and demonstrates the benefit of DDL pre-trained representations for downstream tasks.
            We additionally evaluate motion prediction baselines as pre-training methods for action recognition in Appendix~\ref{appendix:extended-action-recognition}, further highlighting the importance of our framework for representation learning.

\begin{table}[t]
% \vspace{-10pt}
\centering
\caption{\textit{Action recognition results.} 
``Random'' denotes random initialization; 
``Pre-trained'' denotes WorldPose (for WorldPoseAR) or ProSoccer (for SoccerAR) pre-trained model weights.}
\label{tab:action-recognition}
\resizebox{\columnwidth}{!}{%
\begin{tabular}{@{}l|l|c|c@{}}
\toprule
\multirow{2}{*}{Model}
& \multirow{2}{*}{Initialization}
& WorldPoseAR
& SoccerAR \\
&
& Acc. (\%) $\uparrow$
& Acc. (\%) $\uparrow$ \\
\midrule

% Model & Initialization  & Acc. (\%) $\uparrow$ WorldPoseAR & Acc. (\%) $\uparrow$ on SoccerAR  \\ \midrule
ST-GCN~\cite{Yan2018}    & Random                   & \(94.66{\scriptstyle \pm 0.51}\) & \(55.81{\scriptstyle \pm 0.40}\) \\
CTR-GCN~\cite{Chen2021Channel-WiseRecognition}   & Random                   & \(92.42{\scriptstyle \pm 0.44}\) & \(55.31{\scriptstyle \pm 0.70}\) \\
BlockGCN~\cite{Zhou2024BlockGCN:Recognition}  & Random                   & \(78.96{\scriptstyle \pm 4.53}\) & \(57.87{\scriptstyle \pm 1.16}\) \\ 
GTN (ours) & Random                  & \(87.83{\scriptstyle \pm 0.36}\) & \(56.67{\scriptstyle \pm 0.96}\) \\
\midrule
MAMP~\cite{Mao2023MaskedLearners} & Pre-trained (MAMP)       & \(93.17{\scriptstyle \pm 0.50}\) & \(59.12{\scriptstyle \pm 0.18}\) \\
GTN (ours) & Pre-trained (GTN)       & \(95.41{\scriptstyle \pm 0.75}\) & \(62.70{\scriptstyle \pm 1.01}\) \\
GTN (ours) & Pre-trained (GTN+DDL) & \(\textbf{96.21}{\scriptstyle \pm 0.29}\) & \(\textbf{64.15}{\scriptstyle \pm 0.47}\) \\ \bottomrule
\end{tabular}%
}
% \vspace{-10pt}
\end{table}

    % \vspace{-5pt}    
    \subsubsection{Shot Spotting}
    % \vspace{-5pt}
        % \paragraph{Data}
        %     We evaluate shot spotting using an in-house benchmark. Each sample consists of a 50-frame skeleton motion sequence captured by the same player tracking system used for motion prediction. 
        %     Frame-level labels are provided for this task. For positive samples (shots), the shooting frame is annotated as 1 and all other frames as 0, while for negative samples (non-shots) all frames are annotated as 0. 
        %     The benchmark includes 12,191 training samples, 1,991 validation samples, and 3,291 test samples, with positive and negative samples being approximately balanced.

        \textbf{Data.}\quad
            We evaluate shot spotting on an in-house benchmark, where each sample is a 50-frame skeleton motion sequence. Frame-level annotations are provided for all sequences. For positive samples, the shooting frame is labeled 1, and all other frames are labeled 0, whereas for negative samples, all frames are labeled 0.
            The dataset includes over 15,000 samples split into training, validation, and test sets, with approximately balanced positive and negative samples.
            % The dataset includes 12{,}191 training samples, 1{,}991 validation samples, and 3{,}291 test samples, with approximately balanced positive and negative samples.

        \noindent\textbf{Experimental Details.}\quad
            % We fine-tune the GTN model pre-trained on ProSoccer for shot spotting using the same optimization and preprocessing settings as action recognition, except that the batch size is set to 32.
            We fine-tune the GTN model pre-trained on ProSoccer for shot spotting using the same settings as action recognition, except for a batch size of 32. Results are reported as the mean and standard deviation over three random seeds.

        \noindent\textbf{Metric.}\quad
            We evaluate shot spotting performance using Average-AP, following the SoccerNet event spotting protocol~\cite{Giancola2018}. To account for small temporal misalignments, a prediction is considered correct if it falls within a range of $\delta$ frames around the ground-truth shooting frame. For each $\delta$, we compute Average Precision ($\delta$-AP) as the area under the precision-recall curve, and define Average-AP as the mean $\delta$-AP over $\delta \in \{0,1,\ldots,10\}$. Details of Average-AP calculation are provided in Appendix~\ref{appendix:average-ap}.

        \noindent\textbf{Baselines.}\quad
            We use the same baselines as in action recognition, given the limited prior work on skeleton-based event spotting. 
            We set all GCN temporal strides to 1 to preserve sequence length for frame-level prediction. 
            % To enable frame-level prediction, we set the temporal stride to 1 in all GCN layers in the baseline models to preserve the sequence length. 
            Implementation details are provided in Appendix~\ref{appendix:downstream-baseline}.

\begin{table}
% \vspace{-15pt}
\centering
\caption{\textit{Shot spotting results.} ``Random'' denotes random initialization; ``Pre-trained'' denotes ProSoccer pre-trained model weights.}
\label{tab:shot-spotting}
\resizebox{0.82\linewidth}{!}{%
\begin{tabular}{@{}l|l|l@{}}
\toprule
Model    & Initialization              & \multicolumn{1}{c}{Average-AP $\uparrow$} \\ \midrule
ST-GCN \cite{Yan2018}  
& Random                           & \(0.8776{\scriptstyle \pm 0.0070}\)                         \\
CTR-GCN \cite{Chen2021Channel-WiseRecognition}  
& Random                           & \(0.8835{\scriptstyle \pm 0.0010}\)                         \\
BlockGCN \cite{Zhou2024BlockGCN:Recognition} 
& Random                           & \(0.8540{\scriptstyle \pm 0.0324}\)                         \\ 
GTN (ours)     & Random                           & \(0.8957{\scriptstyle \pm 0.0022}\)                          \\
\midrule
MAMP~\cite{Mao2023MaskedLearners}     & Pre-trained (MAMP)           & \(0.8237{\scriptstyle \pm 0.0031}\)                  \\
GTN (ours)     & Pre-trained (GTN)            & \(0.9242{\scriptstyle \pm 0.0035}\)                         \\
GTN (ours)     & Pre-trained (GTN+DDL) & \(\textbf{0.9274}{\scriptstyle \pm 0.0030}\)               \\ \bottomrule
\end{tabular}%
}
% \vspace{-10pt}
\end{table}

        \noindent\textbf{Results.}\quad
            Table~\ref{tab:shot-spotting} presents shot spotting results. 
            As shown, our GTN fine-tuned from GTN+DDL pre-trained weights achieves the best performance, with an Average-AP of 0.9274, significantly outperforming all baselines, including the self-supervised learning method MAMP~\cite{Mao2023MaskedLearners}.
            Its substantial improvement over the randomly initialized GTN confirms the benefits of representation learning. Together with the action recognition results, this demonstrates that the learned representations generalize effectively across multiple downstream tasks. 
            The model initialized from a plain pre-trained GTN remains competitive but shows lower performance, further indicating that DDL helps learn stronger representations.
            We also evaluate motion prediction baselines as pre-training methods for shot spotting in Appendix~\ref{appendix:extended-shot-spotting}, further underscoring the importance of our framework for representation learning beyond motion prediction alone.

% \vspace{-5pt}
\subsection{Ablation Study}
% \vspace{-5pt}
    To analyze the contribution of individual design choices, we conduct a series of ablation studies on both WorldPose and ProSoccer datasets. Unless otherwise stated, all ablation experiments follow the same settings as the motion prediction experiments, except that we set the batch size to 64.

\begin{table}[t]
\centering
% \vspace{-12pt}
\begin{minipage}[t]{0.48\textwidth}
  \centering
  \begin{table}[H]
\centering
\caption{\textit{Ablation study on distribution learning.} ``Reg.'' denotes replacing DDL with regression. The metric is MPJPE $\downarrow$.}
\label{tab:ablation-distribution-worldpose}
\resizebox{\columnwidth}{!}{%
\begin{tabular}{@{}lllllllll@{}}
\toprule
milliseconds    & 80   & 160  & 320  & 400  & 560   & 720   & 880   & 1000  \\ \midrule
\multicolumn{9}{l}{\textit{WorldPose}} \\ \midrule
GTN+DDL &
  \textbf{15.6} &
  \textbf{32.5} &
  \textbf{61.0} &
  \textbf{76.2} &
  \textbf{115.9} &
  \textbf{173.4} &
  \textbf{256.7} &
  338.6 \\
GTN+Reg. & 15.8 & 33.5 & 62.2 & 77.0 & 116.8 & 174.4 & 257.1 & \textbf{338.0} \\
GTN                           & 16.3 & 34.1 & 62.9 & 77.8 & 118.6 & 176.7 & 261.1 & 344.7 \\
\hline\hline
\multicolumn{9}{l}{\textit{ProSoccer}} \\ \midrule
GTN+DDL &
  \textbf{14.1} &
  \textbf{24.9} &
  \textbf{49.6} &
  \textbf{63.0} &
  \textbf{95.2} &
  \textbf{137.1} &
  \textbf{194.1} &
  \textbf{248.0} \\
GTN+Reg. & 15.4 & 28.2 & 54.2 & 67.7 & 101.1 & 144.8 & 203.7 & 259.6 \\
GTN                            & 15.4     & 28.2     &  54.4    &  67.9    & 101.0      &  144.5     &  203.4     & 259.1      \\ 
\bottomrule
\end{tabular}%
}
\end{table}

\end{minipage}\hfill
\begin{minipage}[t]{0.48\textwidth}
  \centering
  \begin{table}[H]
\centering
\caption{\textit{Ablation study on prediction granularity.}
``w/o frame-level'' uses sequence-level supervision instead; ``w/o joint-level'' uses a skeleton-level codebook instead.
The metric is MPJPE $\downarrow$.}
\label{tab:ablation-joint-comparison}

\resizebox{\textwidth}{!}{%
\begin{tabular}{@{}lllllllll@{}}
\toprule
milliseconds & 80 & 160 & 320 & 400 & 560 & 720 & 880 & 1000 \\
\midrule
\multicolumn{9}{l}{\textit{WorldPose}} \\
\midrule
\textbf{Full model} & \textbf{15.6} & \textbf{32.5} & \textbf{61.0} & \textbf{76.2} & \textbf{115.9} & \textbf{173.4} & \textbf{256.7} & \textbf{338.6} \\
w/o frame-level & 18.6 & 40.7 & 78.9 & 97.9 & 144.1 & 212.9 & 311.3 & 404.8 \\
w/o joint-level & 16.3 & 34.0 & 62.9 & 78.1 & 118.4 & 176.3 & 260.0 & 341.6 \\
\hline\hline
\multicolumn{9}{l}{\textit{ProSoccer}} \\
\midrule
\textbf{Full model} & \textbf{14.1} & \textbf{24.9} & \textbf{49.6} & \textbf{63.0} & \textbf{95.2} & \textbf{137.1} & \textbf{194.1} & \textbf{248.0} \\
w/o frame-level & 16.5 & 28.9 & 57.1 & 72.4 & 109.5 & 155.9 & 219.1 & 279.1 \\
w/o joint-level & 15.6 & 28.1 & 53.7 & 67.3 & 101.1 & 145.0 & 204.9 & 262.3 \\
\bottomrule
\end{tabular}
}
\end{table}
\end{minipage}
% \vspace{-10pt}
\end{table}

    \noindent\textbf{Necessity of Distribution Learning.}\quad
        To disentangle the effect of DDL from predicting an intermediate target, we compare our model with a variant that predicts the same future motion via regression. As shown in Table~\ref{tab:ablation-distribution-worldpose}, on both WorldPose and ProSoccer, while direct regression improves over a plain GTN, DDL achieves a further reduction in MPJPE, indicating clear additional gains from modeling a distribution over future motions.

    \noindent\textbf{Necessity of Fine-grained Prediction.}\quad
        Our framework models motion with fine-grained supervision in both time and body structure. 
        Specifically, it predicts future motion at each frame and represents each skeleton motion using $J$ joint-level codes, yielding a discrete motion space of size $K^{J}$ at each frame. 
        We ablate these designs by replacing frame-level supervision with sequence-level supervision, and by replacing joint-level code prediction with a single skeleton-level code supervised by averaged joint motions. 
        As shown in Table~\ref{tab:ablation-joint-comparison}, both variants consistently degrade performance, demonstrating the importance of both frame-level supervision and joint-level codebook design.
        We further show in Appendix~\ref{appendix:ablation-frame-level-application} that frame-level pre-training is crucial for learning representations that transfer effectively to frame-level downstream tasks such as shot spotting.

\begin{figure}
    % \vspace{-15pt} 
    \centering
    \includegraphics[width=\linewidth]{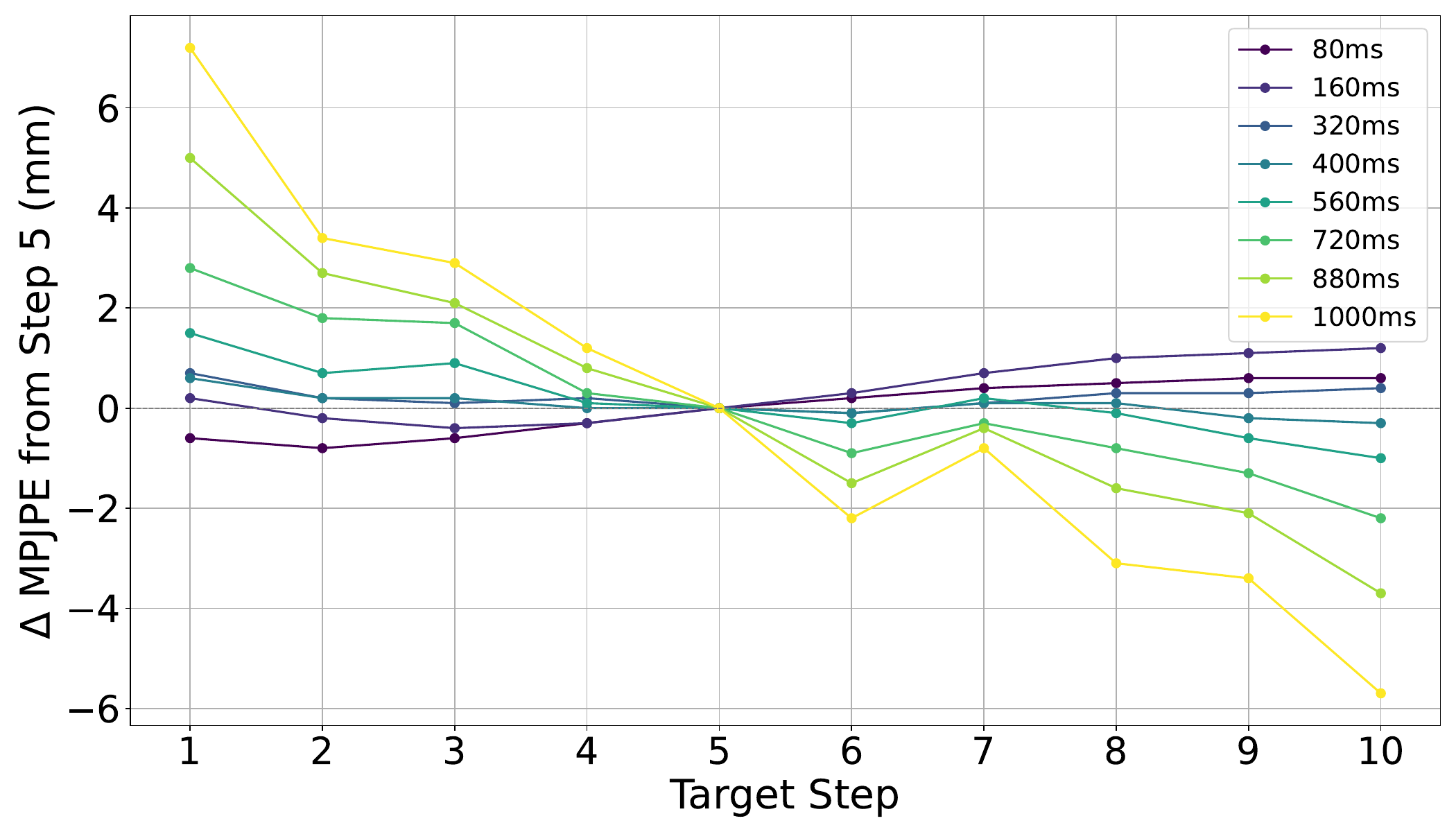}

    \caption{\textit{Impact of the target step on WorldPose.} The plot shows MPJPE changes relative to the default setting (the 5th step), where positive $\Delta$MPJPE indicates worse performance and negative values indicate improvement.}
    \label{fig:step_ablation_worldpose}
    % \vspace{-10pt} 
\end{figure}

    \noindent\textbf{Impact of the Target Step.}\quad
        We study the impact of selecting different future steps as the DDL target.
        Since the prediction horizon is 10 frames,
        we vary the target from the 1st to the 10th future step.
        As shown in Figure~\ref{fig:step_ablation_worldpose} on WorldPose, using earlier steps (e.g., the 2nd step) improves short-term accuracy, yielding lower MPJPE than the default setting (the 5th step), but degrades long-term performance, with MPJPE increasing at horizons beyond 560\,ms. Conversely, using later steps improves long-term prediction at the cost of short-term accuracy.
        Overall, the 5th step provides a balanced trade-off between short- and long-term performance. Similar trends are observed on ProSoccer (Appendix~\ref{appendix:ablation-step}).

    \noindent\textbf{Further Ablation Studies.}\quad
        We conduct further ablation studies in Appendix~\ref{appendix:ablation-codebook} and find that nonlinear code prediction and weight tying consistently improve performance, while sampling during training has a limited impact. 
        We also show in Appendix~\ref{appendix:ablation-hyperparameter} that performance generally improves as the codebook size $K$ increases from 16 to 512, while larger codebooks provide little or no additional benefit.

%% file: sec/5_conclusion.tex
% \vspace{-5pt}
\section{Conclusion and Future Work}
\label{sec:conclusion}
% \vspace{-5pt}

In this work, we exploit motion prediction as a self-supervised objective to learn human motion representations in soccer.
To capture the uncertainty in human motion, we propose learning a discrete distribution over future motion. This is achieved by discretizing the 3D future motion space into a codebook, enabling self-supervised distribution learning over possible motion modes.
Experiments on soccer player tracking data show that our approach significantly improves motion prediction performance and learns representations that transfer effectively to multiple downstream tasks. 
These findings demonstrate the effectiveness of motion prediction with discrete distribution learning for representation learning.
Importantly, our model maintains more stable performance under high-speed motion, highlighting robustness in fast and highly dynamic scenarios.

By focusing on individual player motions, our method provides a foundation for learning player-centric motion representations from large-scale soccer tracking data. 
Beyond individual kinematics, soccer player motion is also shaped by interactions with the ball, teammates, opponents, and the surrounding game context.
Incorporating such factors is a topic for future extensions to enable comprehensive multi-agent and ball-aware representation learning.

%% file: sec/6_acknowledgement.tex
\section*{Acknowledgements}

% WASP
This work was partially supported by the Wallenberg AI, Autonomous Systems and Software Program (WASP) funded by the Knut and Alice Wallenberg Foundation.
% Berzelius
The computations were enabled by the Berzelius resource provided by the Knut and Alice Wallenberg Foundation at the National Supercomputer Centre. 

%% file: sec/X_suppl.tex
\clearpage
\setcounter{page}{1}
% \maketitlesupplementary
\setcounter{section}{0}
\renewcommand{\thesection}{\Alph{section}}

\section{Graph Node and Edge Features}
\label{appendix:graph-construction}
    
    \paragraph{Node Features.}
    Each node corresponds to a joint $j$ (or the player node) at frame $t$, with features encoding its 3D position and joint type (e.g., wrist, elbow, knee).
    For each joint type $t_j$, we learn a trainable embedding $\mathbf{e}_{t_j} \in \mathbb{R}^d$ that captures its semantic identity in a $d$-dimensional latent space,
    while the 3D coordinate $\mathbf{x}_{j,t} \in \mathbb{R}^3$ is linearly projected into the same space. For the player node, we use the pelvis 3D coordinates.
    The node feature is computed as
    \[
    \mathbf{h}_{j,t}^{(0)} = \mathrm{RMSNorm}\bigl( \mathbf{W}_n\mathbf{x}_{j,t} + \mathbf{e}_{t_j} \bigr),
    \]
    where $\mathbf{W}_n \in \mathbb{R}^{d \times 3}$ is a learnable weight matrix, and RMSNorm~\cite{Zhang2019RMSNorm} denotes root mean square normalization. 
    
    \paragraph{Edge Features.}
    Each edge $(v,u)$, defined as directed from $v$ to $u$, carries a feature that encodes the spatial relationship between the two nodes, along with an edge-type indicator specifying whether the connection is skeletal, temporal, or linked to the player node.
    For each edge type $\mathrm{t}(v,u)$, we learn an embedding $\mathbf{e}_{\mathrm{t}(v,u)} \in \mathbb{R}^{d_e}$ in a $d_e$-dimensional space, while the Euclidean distance $\lVert \mathbf{x}_v - \mathbf{x}_u \rVert_2$ is linearly projected into the same latent space.
    The edge feature is computed as
    \[
    \mathbf{a}_{vu} =
    \mathrm{RMSNorm}\bigl(
    \mathbf{W}_e \, \lVert \mathbf{x}_v - \mathbf{x}_u \rVert_2 + \mathbf{e}_{\mathrm{t}(v,u)}
    \bigr),
    \]
    where $\mathbf{W}_e \in \mathbb{R}^{d_e \times 1}$ is learnable and RMSNorm~\cite{Zhang2019RMSNorm} denotes root mean square normalization.

\section{Details of Graph Transformer Network}
\label{appendix:gtn}

\begin{figure}[t]
    \centering
    \includegraphics[width=0.75\columnwidth]{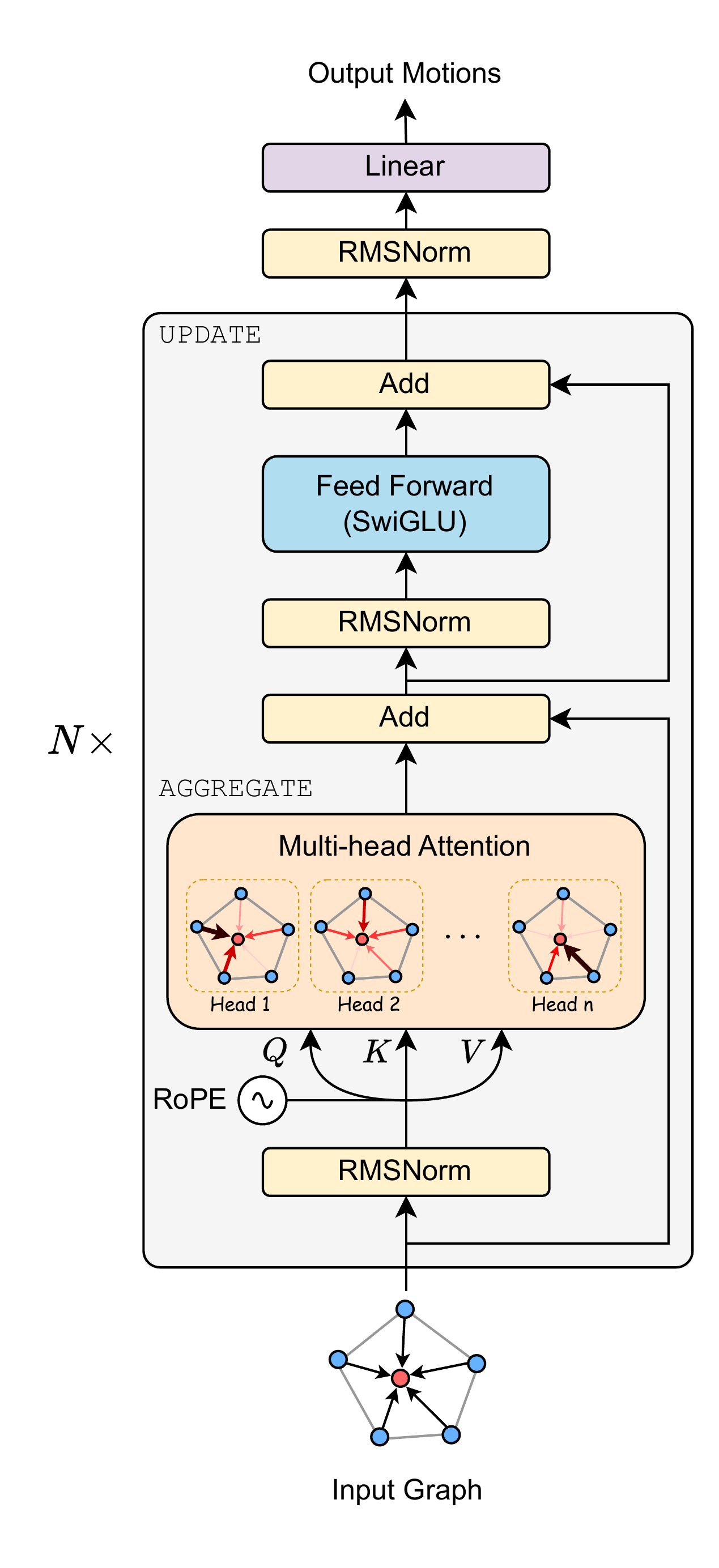}
    % \vspace{-15pt}
    \caption{\textit{Architecture of the Graph Transformer Network (GTN).}}
    \label{fig:gtn}
    % \vspace{-10pt}
\end{figure}

\begin{table*}[t]
\centering
\caption{\textit{Motion prediction results with an additional comparison between GATv2 and GTN.} GATv2 results are highlighted in blue. }
\label{tab:motion-prediction-gatv2}
\resizebox{\textwidth}{!}{%
\begin{tabular}{@{}l|llllllll||llllllll@{}}
\toprule
\multirow{2}{*}{Model}
& \multicolumn{8}{c||}{MPJPE (mm) $\downarrow$ on WorldPose}
& \multicolumn{8}{c}{MPJPE (mm) $\downarrow$ on ProSoccer} \\
\cmidrule(lr){2-9} \cmidrule(lr){10-17}
& 80ms & 160ms & 320ms & 400ms & 560ms & 720ms & 880ms & 1000ms
& 80ms & 160ms & 320ms & 400ms & 560ms & 720ms & 880ms & 1000ms \\
\midrule
Zero %(Repeating last frame)
& 200.4 & 397.3 & 784.6 & 976.5 & 1356.5 & 1734.4 & 2109.4 & 2387.2
& 148.1 & 297.4 & 587.9 & 731.1 & 1015.0 & 1296.9 & 1576.5 & 1783.5 \\
\midrule
LTD \cite{Mao2019LearningPrediction}
& 19.0 & 39.3 & 72.5 & 89.2 & 142.2 & 217.6 & 328.3 & 436.0
& 15.7 & 28.7 & 54.7 & 68.0 & 102.6 & 148.4 & 209.4 & 267.1 \\
HisRep \cite{Mao2020HistoryAttention}
& 18.7 & 39.1 & 72.0 & 88.6 & 132.9 & 195.3 & 287.4 & 376.3
& 15.3 & 28.3 & 54.1 & 67.4 & 100.2 & 143.7 & 202.2 & 257.4 \\
MSR-GCN \cite{Dang2021MSR-GCN:Prediction}
& 23.7 & 46.5 & 81.3 & 100.3 & 161.2 & 238.5 & 351.5 & 459.4
& 24.9 & 43.6 & 76.6 & 94.6 & 144.2 & 205.0 & 289.9 & 362.9 \\
PGBIG \cite{Ma2022ProgressivelyPrediction}
& 17.3 & 37.8 & 69.0 & 85.6 & 135.1 & 202.7 & 305.6 & 405.8
& 17.8 & 34.1 & 64.7 & 80.7 & 121.6 & 170.7 & 238.2 & 300.9 \\
SiMLPe \cite{Guo2023BackPrediction}
& 19.7 & 43.7 & 83.5 & 103.5 & 158.2 & 232.7 & 341.2 & 445.5
& 17.7 & 34.3 & 66.6 & 83.4 & 124.7 & 176.7 & 245.0 & 308.7 \\
GCNext \cite{Wang2024GCNext:Prediction}
& 18.5 & 41.3 & 77.5 & 95.4 & 142.7 & 211.1 & 308.2 & 403.2
& 18.8 & 37.0 & 71.2 & 88.6 & 130.8 & 184.3 & 253.8 & 317.8 \\
\midrule
\rowcolor{blue!8}
GATv2 \cite{Brody2021HowNetworks}
& 26.5 & 48.6 & 83.2 & 102.5 & 155.6 & 231.3 & 341.1 & 447.4
& 17.0 & 30.1 & 56.9 & 71.0 & 106.0 & 151.1 & 211.7 & 268.9 \\
\midrule
GTN (ours)
& \textbf{16.3} & \textbf{34.1} & \textbf{62.9} & \textbf{77.8} & \textbf{118.6} & \textbf{176.7} & \textbf{261.1} & \textbf{344.7}
& \textbf{15.2} & \textbf{27.8} & \textbf{53.2} & \textbf{66.4} & \textbf{99.0} & \textbf{141.8} & \textbf{199.6} & \textbf{254.7} \\
\bottomrule
\end{tabular}
}
\end{table*}

\subsection{Model Architecture}

In this section, we introduce our Graph Transformer Network (GTN) as a GNN backbone for this work. A GNN updates node representations by iteratively aggregating information from neighboring nodes.
For a node $u$, let its incoming neighbors be
$\mathcal{N}_{u} = \{ v \mid (v, u) \in \mathcal{E} \}$,
where $\mathcal{E}$ is the edge set and $(v, u)$ is a directed edge from node $v$ to node $u$.
Each node $u$ has a node feature $\mathbf{h}_u^{(0)}$, and each edge $(v, u)$ has an edge feature $\mathbf{a}_{vu}$.

Let $\mathbf{h}_{u}^{(k)}$ denote the representation of node $u$ at layer $k$.
The general message-passing for one GNN layer is given by
% \begin{equation}
% \label{eqn:gnn}
% % \scriptsize
%     \mathbf{h}_{u}^{(k)} = \text{UPDATE} (\mathbf{h}_{u}^{(k-1)}, \text{AGGREGATE}(\mathbf{h}_{u}^{(k-1)}, \{(\mathbf{h}_{v}^{(k-1)}, \mathbf{e}_{vu}) | v \in \mathcal{N}_u\})).
% \end{equation}
\begin{equation}
\label{eqn:gnn}
\begin{aligned}
\mathcal{M}_{u}^{(k-1)}
&=
\{(\mathbf{h}_{v}^{(k-1)}, \mathbf{a}_{vu})
  \mid v \in \mathcal{N}_{u}\},
\\
\mathbf{m}_{u}^{(k)}
&=
\operatorname{AGGREGATE}\bigl(
    \mathbf{h}_{u}^{(k-1)},
    \mathcal{M}_{u}^{(k-1)}
\bigr),
\\
\mathbf{h}_{u}^{(k)}
&=
\operatorname{UPDATE}\bigl(
    \mathbf{h}_{u}^{(k-1)},
    \mathbf{m}_{u}^{(k)}
\bigr),
\end{aligned}
\end{equation}
% where $\mathrm{AGGREGATE}(\cdot)$ denotes any permutation-invariant aggregation function, and $\mathrm{UPDATE}(\cdot)$ is a learnable function that updates the node representation based on the aggregated information and its previous state.
where \(\mathcal{M}{u}^{(k-1)}\) denotes the set of neighboring node representations and their associated edge features, \(\mathrm{AGGREGATE}(\cdot)\) is any permutation-invariant function that produces the aggregated message \(\mathbf{m}_{u}^{(k)}\), and \(\mathrm{UPDATE}(\cdot)\) is a learnable function that updates the node representation by combining the aggregated message with its previous state.

We follow modern Transformer designs to implement GTN, including pre-normalization~\cite{Xiong2020OnLayerNormalization} with RMSNorm~\cite{Zhang2019RMSNorm}, an FFN with SwiGLU activation~\cite{Shazeer2020GLUTransformer}, and rotary positional embeddings (RoPE)~\cite{Su2024RoFormer:Embedding} to encode temporal information.
The aggregation function $\mathrm{AGGREGATE}(\cdot)$ is implemented using multi-head attention. An additional RMSNorm is applied after the final layer. An overview of the GTN architecture is shown in Figure~\ref{fig:gtn}.

At layer $k$, pre-normalization is first applied to node representations:
\begin{equation}
\begin{aligned}
\tilde{\mathbf{h}}_{u}^{(k-1)}
&= \mathrm{RMSNorm}\!\left(\mathbf{h}_{u}^{(k-1)}\right), \\
\tilde{\mathbf{h}}_{v}^{(k-1)}
&= \mathrm{RMSNorm}\!\left(\mathbf{h}_{v}^{(k-1)}\right).
\end{aligned}
\end{equation}

The attention weight from node $v$ to node $u$ is then computed as
% \begin{equation}
% % \small
% \alpha_{uv} =
% \mathrm{softmax}_{v}\!\left(
% \frac{
% \mathrm{RoPE}( \mathbf{W}_Q \tilde{\mathbf{h}}_{u}^{(k-1)} )
% \left(
% \mathrm{RoPE}( \mathbf{W}_K \tilde{\mathbf{h}}_{v}^{(k-1)} ) + \mathbf{e}_{vu}
% \right)^{\top}
% }{\sqrt{d_k}}
% \right),
% \end{equation}
\begin{equation}
\begin{aligned}
\mathbf{q}_{u}
&=
\operatorname{RoPE}\!\left(
\mathbf{W}_Q \tilde{\mathbf{h}}_{u}^{(k-1)}
\right), \\
\mathbf{k}_{v}
&=
\operatorname{RoPE}\!\left(
\mathbf{W}_K \tilde{\mathbf{h}}_{v}^{(k-1)}
\right)
+\mathbf{a}_{vu}, \\
\alpha_{uv}
&=
\operatorname{softmax}_{v \in \mathcal{N}_u}
\left(
\frac{\mathbf{q}_{u}^{\top}\mathbf{k}_{v}}
{\sqrt{d_k}}
\right),
\end{aligned}
\end{equation}
% where 
% $\mathbf{W}_Q$ and $\mathbf{W}_K$ are learnable projections, 
% $d_k$ denotes the dimensionality of the key vectors, and $\mathrm{RoPE}$ refers to rotary positional encoding \cite{Su2024RoFormer:Embedding} with the temporal index of the node.
where \(\mathbf{q}_{u}\) and \(\mathbf{k}_{v}\) denote the query and key vectors, respectively; \(\mathbf{W}_Q\) and \(\mathbf{W}_K\) are learnable projection matrices; \(d_k\) is the dimensionality of the key vectors; and \(\operatorname{RoPE}(\cdot)\) denotes rotary positional encoding \cite{Su2024RoFormer:Embedding}, applied using the temporal index of the corresponding node.

The $\mathrm{AGGREGATE}(\cdot)$ function computes a weighted sum over neighbors, followed by a linear projection:
\begin{equation}
\mathrm{agg}_{u}^{(k-1)} =
\mathbf{W}_O
\sum_{v \in \mathcal{N}_u}
\alpha_{uv}\,
\mathbf{W}_V \tilde{\mathbf{h}}_{v}^{(k-1)},
\end{equation}
where $\mathbf{W}_V$ and $\mathbf{W}_O$ are learnable projections.

For the update function $\mathrm{UPDATE}(\cdot)$, we apply a feed-forward network (FFN) with residual connections:
% \begin{equation}
% \label{eqn:gtn_update}
% \begin{aligned}
% \mathbf{h}_{u}^{(k)}
% &= \mathrm{UPDATE}\!\left(\mathbf{h}_{u}^{(k-1)}, \mathrm{agg}_{u}^{(k-1)}\right) \\
% &= \mathbf{h}_{u}^{(k-1)} +
% \mathrm{FFN}\!\left(
% \mathrm{RMSNorm}\!\left(\mathbf{h}_{u}^{(k-1)} +\mathrm{agg}_{u}^{(k-1)}\right)
% \right).
% \end{aligned}
% \end{equation}
\begin{equation}
\label{eqn:gtn_update}
\begin{aligned}
\mathbf{z}_{u}^{(k-1)}
&=
\mathbf{h}_{u}^{(k-1)}
+\mathrm{agg}_{u}^{(k-1)}, \\
\mathbf{h}_{u}^{(k)}
&=
\operatorname{UPDATE}\!\left(
\mathbf{h}_{u}^{(k-1)},
\mathrm{agg}_{u}^{(k-1)}
\right) \\
&=
\mathbf{z}_{u}^{(k-1)}
+\operatorname{FFN}\!\left(
\operatorname{RMSNorm}\!\left(
\mathbf{z}_{u}^{(k-1)}
\right)
\right),
\end{aligned}
\end{equation}
% where $\mathrm{FFN}(\cdot)$ denotes a position-wise feed-forward network with SwiGLU activation.
where \(\mathbf{z}_{u}^{(k-1)}\) denotes the residual combination of the previous node representation and the aggregated message, and \(\operatorname{FFN}(\cdot)\) is a position-wise feed-forward network with a SwiGLU activation.

\subsection{GTN for Motion Prediction}
\label{appendix:gtn-mse-loss}

Let $\mathbf{h}_{j,t}^{(L)} \in \mathbb{R}^{d}$ denote the final-layer representation of joint $j$ at frame $t$ from an $L$-layer GTN. A shared linear layer is then applied to predict the next $T$-frame motion of this joint $\hat{\bm{m}}_{j,t+1:t+T} \in \mathbb{R}^{T \times 3}$:
\begin{equation}
\label{eqn:linear_layer_motion_prediction}
    \begin{aligned}
    \hat{\bm{m}}_{j,t+1:t+T}
    =
    \mathbf{W}_p \mathbf{h}_{j,t}^{(L)},
    \end{aligned}
\end{equation}
where $\mathbf{W}_p$ is a learnable projection from $\mathbb{R}^d$ to $\mathbb{R}^{T \times 3}$.

For skeletons with $J$ joints, the model is trained with an MSE loss between the predicted future motion
$\hat{\bm{M}}_{t+1:t+T} = \{\hat{\bm{m}}_{j,t+1:t+T}\}_{j=1}^{J}$ and the ground-truth future motion $\bm{M}_{t+1:t+T} = \{\mathbf{m}_{j, t+1:t+T} \}_{j=1}^{J}$, where $\mathbf{m}_{j, t+\tau} = \mathbf{x}_{j, t+\tau} - \mathbf{x}_{j, t}$:
\begin{equation}
\label{eqn:motion_loss}
\small
\mathcal{L}_{\mathrm{motion}}
=
\frac{1}{N}
\sum_{t=1}^{N}
\frac{1}{J T}
\sum_{j=1}^{J}
\sum_{\tau=1}^{T}
\left\|
\hat{\bm{m}}_{j,t+\tau}
-
\bm{m}_{j,t+\tau}
\right\|_2^2 .
\end{equation}

\subsection{Empirical Comparison of GTN and GATv2}
    Our GTN adopts multi-head attention to aggregate information from neighboring nodes.
    This shares the attention-based message passing paradigm with GATv2~\cite{Brody2021HowNetworks}, although GATv2 employs a linear attention formulation.
    % similar to GATv2~\cite{Brody2021HowNetworks}, which also relies on attention-based message passing. 
    To evaluate the impact of transformer-like attention in our GTN, we conduct motion prediction experiments with GATv2 as the baseline and compare GTN with GATv2 on both the WorldPose and ProSoccer datasets.

    We train GATv2 under the same settings as GTN on the WorldPose and ProSoccer datasets. 
    % To ensure a fair comparison, we use identical architectural configurations: an 8-layer GATv2 with 512 hidden dimensions for WorldPose, and a 12-layer GATv2 with 1024 hidden dimensions for ProSoccer. 
    To ensure a fair comparison, we use identical architectural configurations: an 8-layer GATv2 with a hidden dimension of 512 for WorldPose and a 12-layer GATv2 with a hidden dimension of 1024 for ProSoccer.
    Motion sequences are centered at the pelvis of the first frame, with random $xy$-plane rotations applied as data augmentation during training.
    Training follows the same protocol as GTN. We use AdamW as the optimizer with a weight decay of 0.1, along with a cosine annealing scheduler with 2\% linear warm-up. Models on WorldPose are trained for 50k steps with a learning rate of $2\times10^{-4}$ and a batch size of 64 on a single NVIDIA H200 GPU. Models on ProSoccer are trained for 100k steps with a learning rate of $3\times10^{-4}$ and a batch size of 256 across four NVIDIA H200 GPUs.

    Table \ref{tab:motion-prediction-gatv2} presents the results of the GATv2 model (highlighted in blue), along with the results previously reported in Table \ref{tab:motion-prediction-merged}.
    Across both WorldPose and ProSoccer, GTN consistently and significantly outperforms GATv2 across all prediction horizons. This performance gap suggests that, in addition to attention-based message passing, the architectural components introduced in GTN, including multi-head attention and nonlinear feedforward transformations, provide greater representational capacity for modeling complex motion dynamics. These findings support the choice of GTN as the backbone model in this work.

\section{Balanced KD-tree Motion Codebook}
\label{appendix:kd-tree-construction}

Here we provide the algorithmic details of the balanced KD-tree used to construct the motion codebook. We construct the KD-tree from future motion samples extracted from the training set.

Formally, for a joint $j$ at frame $t$, the codebook is constructed from motions at a fixed future offset of $t_f$ frames. A future motion sample is defined as
\begin{equation}
\bm{m}_{j,t}
=
\bm{x}_{j,t+t_f}
-
\bm{x}_{j,t}
\in
\mathbb{R}^{3},
\end{equation}
where $\bm{x}_{j,t} \in \mathbb{R}^{3}$ denotes the 3D coordinates of joint $j$ at frame $t$.
We collect such motions $\mathcal{M}=\{\bm{m}_i \in \mathbb{R}^{3}\}_{i=1}^{|\mathcal{M}|}$ from the training set and partition the continuous 3D motion space into $K$ regions using a balanced KD-tree algorithm, as illustrated in Algorithm~\ref{alg:kd_codebook_build}.
The algorithm recursively splits the motion space along the dimension with the highest variance using median thresholds, producing balanced partitions with comparable sample counts.
This balanced partitioning ensures near-uniform code utilization, enabling effective learning of multimodal future motion distributions.

\begin{algorithm}[t]
        \caption{\small Balanced KD-tree Motion Codebook Construction}
        \label{alg:kd_codebook_build}
        \begin{algorithmic}
        \STATE {\bfseries Input:} motion samples $\mathcal{M}=\{\bm{m}_i \in \mathbb{R}^{3}\}_{i=1}^{|\mathcal{M}|}$, $\bm{m}_i = (m_i^{x},m_i^{y},m_i^{z})$; target codebook size $K$
        \STATE {\bfseries Output:} KD-tree $\mathcal{T}$ with $K$ leaf nodes, each leaf corresponds to a region in the 3D motion space
        \newline
        % \STATE Each node $u$ stores an associated sample set $\mathcal{S}(u)$
        \STATE Create root node $r$ with all samples: $\mathcal{S}(r)\gets \mathcal{M}$, where $\mathcal{S}(r)$ denotes the sample set associated with node $r$
        \STATE Initialize leaf set $\mathcal{L}\gets \{r\}$
        
        \WHILE{$|\mathcal{L}| < K$}
            \STATE Select the leaf with the most samples:
            \[
                u \gets \arg\max_{v\in\mathcal{L}} |\mathcal{S}(v)|
            \]
            \STATE Select the split axis with maximum variance:
            \[
            a_u \gets \mathrm{argmax}_{d \in \{x,y,z\}} \mathrm{Var}\!\left(\{m_i^{d} : \bm{m}_i\in\mathcal{S}(u)\}\right)
            \]
            \STATE Set the split threshold as the median along axis $a_u$:
            \[
            \theta_u \gets \mathrm{Median}\big(\{ m_i^{a_u} : \bm{m}_i \in \mathcal{S}(u) \}\big)
            \]
            \STATE Create children nodes $u_L,u_R$ and partition samples:
            \[
            \mathcal{S}(u_L) \gets \{\bm{m}_i \in \mathcal{S}(u) \mid m_i^{a_u} \le \theta_u\}
            \]
            \[
            \mathcal{S}(u_R) \gets \{\bm{m}_i \in \mathcal{S}(u) \mid m_i^{a_u} > \theta_u\}
            \]
            \STATE If $|\mathcal{S}(u_L)| = 0$ \textbf{or} $|\mathcal{S}(u_R)| = 0$, \textbf{break}
            \STATE Update leaf set $\mathcal{L} \leftarrow (\mathcal{L} \setminus \{u\}) \cup \{u_L, u_R\}$
        \ENDWHILE
        
        \STATE Assign code indices $\{1,\ldots,K\}$ to leaf nodes in $\mathcal{L}$
        \STATE \textbf{Return} $\mathcal{T}$, rooted at $r$ with stored $\{a_u,\theta_u,u_L,u_R$\} for each internal node $u$
        \end{algorithmic}
    \end{algorithm}

\section{Experimental Details}
\label{appendix:exp_details}
    \subsection{Experimental Details for Motion Prediction}
    \label{appendix:exp_details_motion_prediction}

    We use different configurations for WorldPose and ProSoccer. 
    % For WorldPose, we employ an 8-layer GTN with 512 hidden dimensions and 8 attention heads, while for ProSoccer we use a 12-layer GTN with 1024 hidden dimensions and 16 heads. 
    For WorldPose, we use an 8-layer GTN with a hidden dimension of 512 and 8 attention heads. For ProSoccer, we use a 12-layer GTN with a hidden dimension of 1024 and 16 attention heads.
    The codebook size is set to 512, constructed using motion at the 5th future frame ($t_f = 5$). The loss weight $\alpha$ is set to 0.001 for WorldPose and 0.01 for ProSoccer. Motion sequences are centered at the pelvis of the first frame, with random $xy$-plane rotations applied as data augmentation during training.
    
    All models are trained using AdamW~\cite{loshchilov2018decoupled} with a weight decay of 0.1 and a cosine annealing scheduler with a 2\% linear warm-up. WorldPose models are trained for 50k steps with a learning rate of $2\times10^{-4}$ and a batch size of 64 on a single NVIDIA H200 GPU.
    % , taking approximately 12 hours. 
    ProSoccer models are trained for 100k steps with a learning rate of $3\times10^{-4}$ and a batch size of 256 across four NVIDIA H200 GPUs.
    % , taking approximately 50 hours.

    \subsection{Experimental Details for Action Recognition}
    \label{appendix:ar_details}

    We fine-tune a WorldPose pre-trained GTN for WorldPoseAR and a ProSoccer pre-trained GTN for SoccerAR. We train for ten epochs with a learning rate of $2\times10^{-4}$, using AdamW with a weight decay of 0.1, and a cosine annealing scheduler with 1000 warm-up steps. We follow the same preprocessing as in motion prediction, except without random $xy$-plane rotations. Training uses a batch size of 16 on a single NVIDIA H200 GPU.
    % and takes approximately one hour.
    % We report classification accuracy on the test set.

\begin{table*}[t]
\centering
\caption{\textit{Motion prediction baseline model settings.}}
\label{tab:baseline-motion-prediction-merged}
\resizebox{\textwidth}{!}{%
\begin{tabular}{@{}l|llllll||llllll@{}}
\toprule
\multirow{2}{*}{Setting}
& \multicolumn{6}{c||}{WorldPose}
& \multicolumn{6}{c}{ProSoccer} \\
\cmidrule(lr){2-7} \cmidrule(lr){8-13}
& LTD & HisRep & MSR-GCN & PGBIG & SiMLPe & GCNext
& LTD & HisRep & MSR-GCN & PGBIG & SiMLPe & GCNext \\
\midrule
Number of Layers / Stages
& 12 & 12 & 4 & 12 & 48 & 48
& 24 & 24 & 4 & 24 & 48 & 48 \\
Hidden Dimension
& 256 & 256 & 512 & 32 & 63 & 63
& 512 & 512 & 512 & 32 & 63 & 63 \\
Dropout Rate
& 0.2 & 0.1 & 0.1 & 0.1 & 0 & 0
& 0 & 0 & 0.1 & 0.1 & 0 & 0 \\
Training Steps
& 800k & 800k & 200k & 800k & 200k & 200k
& 800k & 800k & 800k & 800k & 200k & 200k \\
Batch Size
& 64 & 64 & 64 & 64 & 256 & 256
& 64 & 64 & 64 & 64 & 256 & 256 \\
Learning Rate
& 2e-4 & 2e-4 & 2e-4 & 2e-4 & 5e-4 & 6e-4
& 1e-4 & 1e-4 & 2e-4 & 2e-4 & 5e-4 & 5e-4 \\
DCT Length
& 50 & 20 & 35 & 20 & 50 & 50
& 50 & 20 & 35 & 20 & 50 & 50 \\
Auxiliary Learning Rate
&  &  &  &  &  & 1e-8
&  &  &  &  &  & 1e-8 \\
\bottomrule
\end{tabular}
}
\end{table*}

\section{Details of Baseline Methods}
    \subsection{Details of Motion Prediction Baselines}
    \label{appendix:motion-prediction-baseline}

We adopt 6 prior works as baselines for motion prediction. 
        
\noindent\textbf{LTD} \cite{Mao2019LearningPrediction} introduces motion prediction in the trajectory space using a discrete cosine transform (DCT) together with a learnable graph structure to model spatial dependencies.

\noindent\textbf{HisRep} \cite{Mao2020HistoryAttention} proposes to use attention over observations to capture repeating motion patterns.

\noindent\textbf{MSR-GCN} \cite{Dang2021MSR-GCN:Prediction} proposes a multi-scale fine-to-coarse-to-fine architecture with explicit supervision at each resolution level.

\noindent\textbf{PGBIG} \cite{Ma2022ProgressivelyPrediction}  progressively refines initial guesses for motion prediction using recursively smoothed ground-truth sequences as supervision.

\noindent\textbf{SiMLPe} \cite{Guo2023BackPrediction} demonstrates that a lightweight MLP with standard training practices can achieve competitive motion prediction performance.

\noindent\textbf{GCNext} \cite{Wang2024GCNext:Prediction} proposes a universal graph convolution framework that dynamically selects graph convolution operations across layers and samples.

As our focus is on representation learning, we do not include diffusion-based or flow-based methods as baselines. Unlike the predictive baselines listed above, diffusion-based or flow-based models do not provide a consistent per-sample latent representation that can be adapted for downstream tasks, making them unsuitable for representation learning purposes.

All models are trained using the AdamW optimizer with a weight decay of \(1\times10^{-4}\). A cosine annealing learning rate scheduler is employed, with the first 1{,}000 training steps used for warm-up. Skeleton sequences are centered at the pelvis joint in the first frame, and random rotation augmentation in the \(xy\)-plane is applied during training. Hyperparameters used for baseline models are shown in Table \ref{tab:baseline-motion-prediction-merged}.

    \subsection{Details of Baselines for Downstream Tasks}
    \label{appendix:downstream-baseline}

        We adopt three representative supervised learning methods and one self-supervised learning method as baseline methods for both action recognition and shot spotting.

        \subsubsection{Supervised Learning Baselines}
            
            \noindent\textbf{ST-GCN}~\cite{Yan2018} is a classic baseline that models human skeletons as spatio-temporal graphs and applies graph convolutions over joints and time.
            
            \noindent\textbf{CTR-GCN}~\cite{Chen2021Channel-WiseRecognition} introduces channel-wise topology refinement to adaptively learn joint dependencies.
            
            \noindent\textbf{BlockGCN}~\cite{Zhou2024BlockGCN:Recognition} preserves skeletal topology by encoding physical joint connectivity and employs block-based graph convolution.

        For shot spotting, we modify the models by setting the temporal stride to 1 across all GCN layers so that the output sequence length matches the input sequence length. We then apply a linear head followed by a sigmoid function to the per-frame representations to produce frame-level confidence scores.

        All three models are trained using the AdamW optimizer with a weight decay of $1 \times 10^{-4}$, and a cosine annealing scheduler with the first 1000 steps reserved for linear warm-up. 
        % Motion sequences are preprocessed to be centered at the pelvis in the first frame. We do not use random rotation augmentation, as player orientation may provide useful cues for these tasks.
        We use 10 layers and a hidden dimension of 256 for these models. For action recognition, we train them for 50 epochs with a batch size of 16 and a learning rate of $3\times10^{-4}$; the dropout rate is set to 0.5 for ST-GCN and BlockGCN, and 0.2 for CTR-GCN. For shot spotting, we also train for 50 epochs with a batch size of 32 and a dropout rate of 0.5. The learning rate is set to $5\times10^{-3}$ for ST-GCN and CTR-GCN, and $2\times10^{-3}$ for BlockGCN.

        \subsubsection{Self-supervised Learning Baseline}

        \noindent\textbf{MAMP}~\cite{Mao2023MaskedLearners} is a transformer-based self-supervised method that learns skeleton representations by masking motion-rich regions and predicting the masked joint motion.

        We pre-train two MAMP models separately on WorldPose and ProSoccer. The WorldPose pre-trained model is fine-tuned for action recognition on WorldPoseAR, while the ProSoccer pre-trained model is fine-tuned for action recognition on SoccerAR and for shot spotting.      
        % We pre-train MAMP on the corresponding unlabeled training set: WorldPose for WorldPoseAR, and ProSoccer for SoccerAR and shot spotting. 
        We use a spatial patch size of one joint and a temporal patch size of five frames, corresponding to 0.2 seconds. We mask 90\% of the spatio-temporal tokens using motion-aware masking with $\tau=0.8$ and predict normalized joint motion. The encoder consists of eight transformer layers with a hidden dimension of 256 and eight attention heads, while the decoder consists of three layers with a hidden dimension of 256. We pre-train MAMP for 100,000 steps with a batch size of 32 using AdamW with a weight decay of 0.05. The learning rate follows a cosine schedule from $3\times10^{-4}$ to $1\times10^{-5}$ after 5,000 linear warm-up steps. Random $xy$-plane rotation augmentation is applied only during pre-training.
        
        For action recognition, we average the encoder representations over the temporal dimension, concatenate the representations of joints, and pass them through a classification MLP with a hidden dimension of 2,048 and a dropout rate of 0.3. The model is optimized using cross-entropy loss with label smoothing of 0.1.
        
        For shot spotting, we average the encoder representations over the joint dimension and apply a linear head to each temporal-patch representation. The head predicts five frame-level logits for each patch, which are reshaped to recover the original 50-frame temporal resolution. The logits are optimized using binary cross-entropy loss, and a sigmoid function is applied at inference time to obtain frame-level confidence scores.

        For both tasks, we fine-tune the pre-trained MAMP encoder for 10 epochs using AdamW with a learning rate of $2\times10^{-4}$, a weight decay of 0.1, and a cosine schedule with 1,000 linear warm-up steps. The batch sizes are 16 for action recognition and 32 for shot spotting.

\section{Details of Action Recognition Benchmarks}
\label{sec:appendix_ar_benchmarks}

\subsection{WorldPoseAR}

WorldPoseAR is a soccer action recognition benchmark constructed using skeletal sequences from the WorldPose dataset~\cite{Jiang2024WorldPose:Estimation}. It comprises 8 action classes: \texttt{STANDING}, \texttt{WALKING}, \texttt{JOGGING}, \texttt{RUNNING}, \texttt{SIDE\_STEPPING}, \texttt{JOGGING\_BACK}, \texttt{WALKING\_BACK}, and \texttt{KICKING}. The \texttt{KICKING} class includes all ball-kicking actions, such as short passing, long passing, crossing, and shooting, while the remaining seven classes correspond to off-ball player movement actions.

The dataset is split into training, validation, and test sets with 3,058, 400, and 800 samples, respectively. The validation and test sets are class-balanced, containing 50 and 100 samples per class. Each sample comprises a 50-frame skeleton sequence at 25 Hz, with a single class label.

\subsection{SoccerAR}

SoccerAR is an in-house benchmark for soccer action recognition, comprising 35 classes and over 20,000 samples. It covers a wide range of actions on the pitch, including both on-ball actions (e.g., shooting, passing, dribbling) and off-ball actions (e.g., running, walking, bending). The dataset is split into training, validation, and test sets, with the validation and test sets containing 50 samples per class. The skeletal sequences in this dataset are derived from the same tracking data as ProSoccer. 
% Each class represents a distinct player action. 
% Table~\ref{tab:action_classes} lists all action classes.

Each sample consists of a 50-frame skeleton sequence annotated with a class label.
Although some action classes are not mutually exclusive in principle, each sample in the dataset is annotated with exactly one class label.
When multiple actions occur simultaneously, on-ball actions are prioritized.
For example, if a player is running while shooting, the sample is labeled as \texttt{SHOOTING}.

\begin{figure*}[t]
    \centering

    \begin{minipage}[t]{0.48\textwidth}
        \centering
        \includegraphics[width=\linewidth]
        {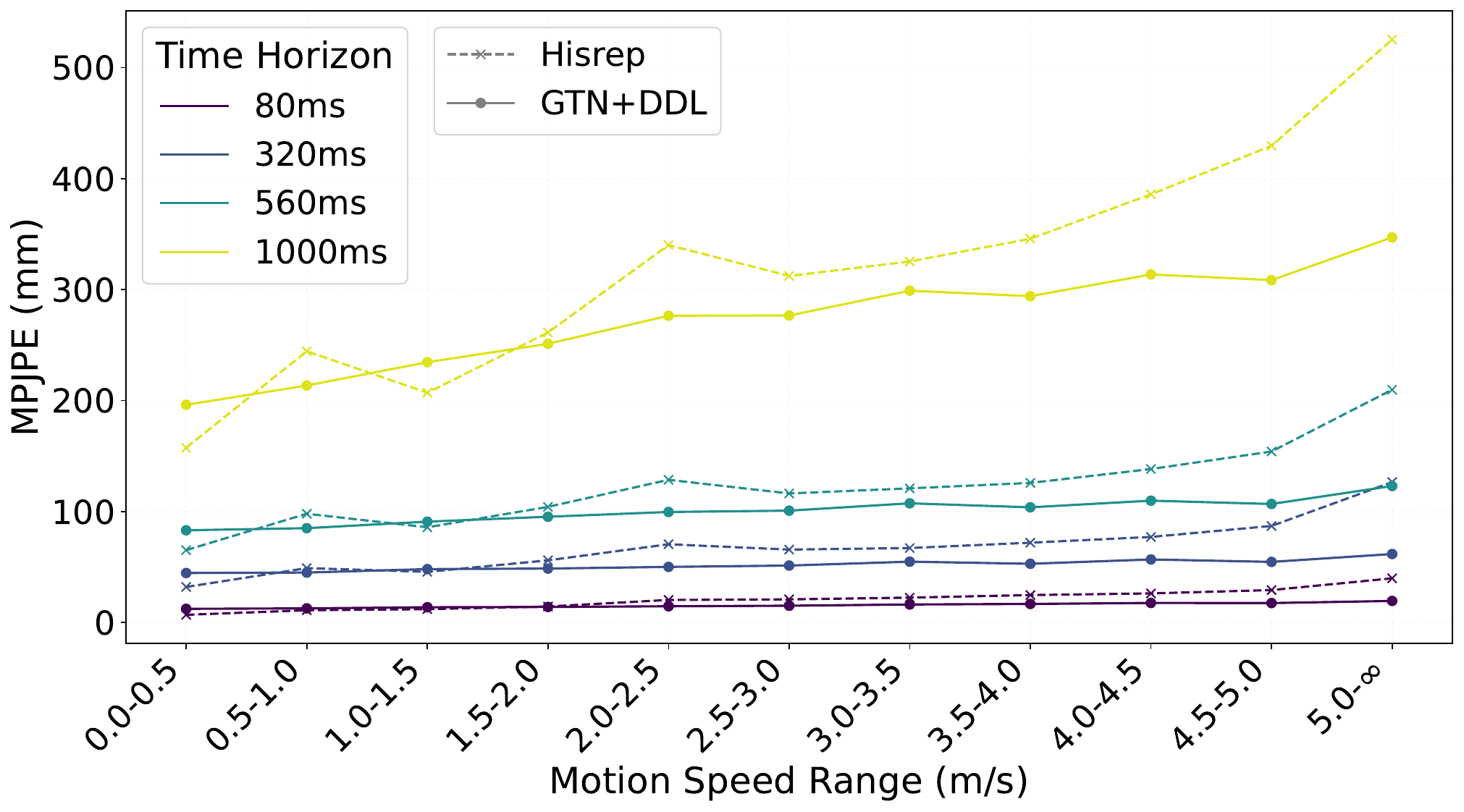}
        \captionof{figure}{
        \textit{Motion prediction performance across different motion
        speeds on ProSoccer.}
        We compare HisRep (dashed) and our GTN+DDL model (solid).
        }
        \label{fig:range_eval_ProSoccer}
    \end{minipage}
    \hfill
    \begin{minipage}[t]{0.48\textwidth}
        \centering
        \includegraphics[width=\linewidth]
        {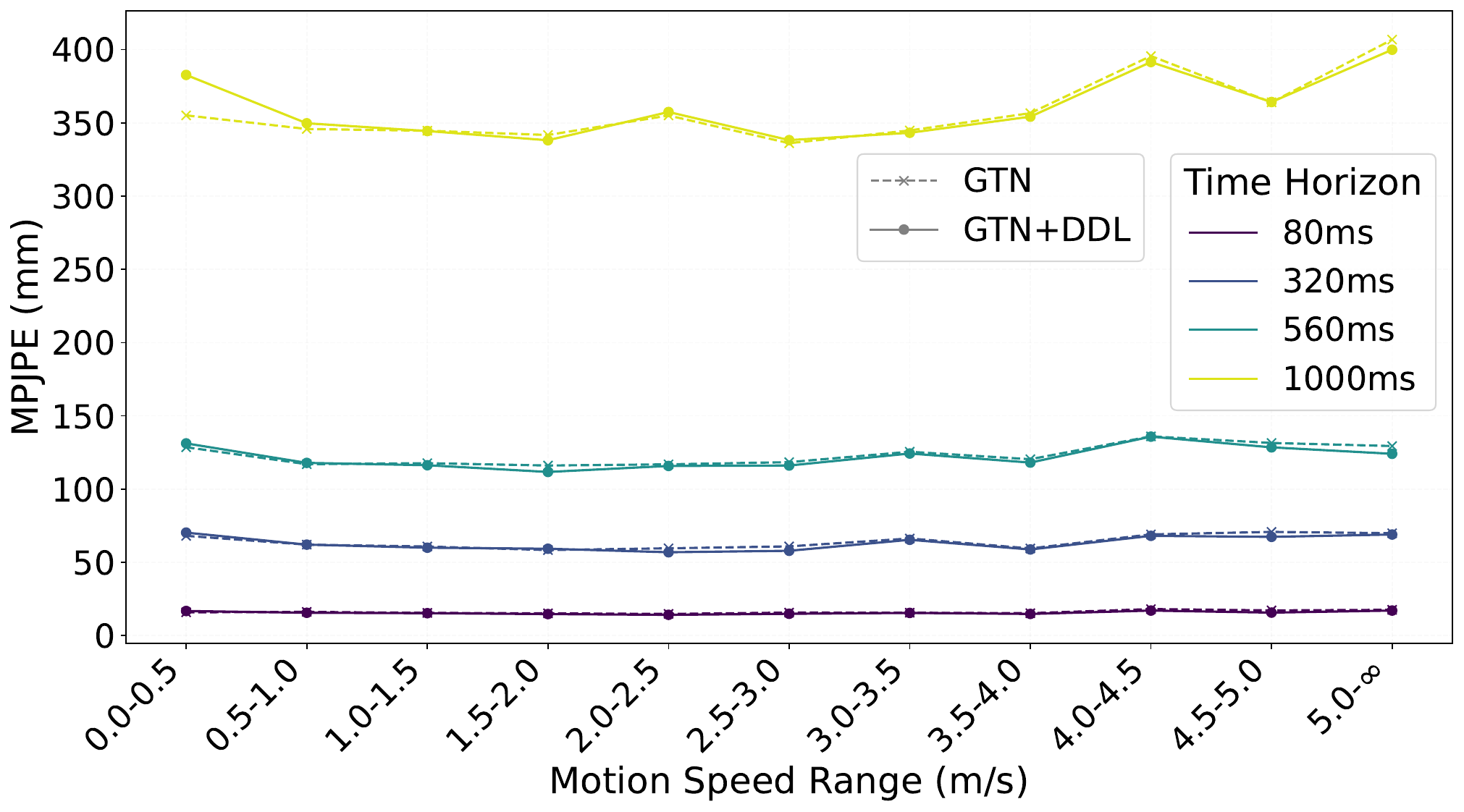}
        \captionof{figure}{
        \textit{Motion prediction performance across different motion
        speeds on WorldPose.}
        We compare GTN with DDL (solid) and without DDL (dashed).
        }
        \label{fig:range_eval_worldpose_gtn}
    \end{minipage}
% \vspace{-5pt}
\end{figure*}

\section{Details of
\texorpdfstring{$\mathrm{Average\text{-}AP}$}{Average-AP}
Calculation}
\label{appendix:average-ap}

    We adopt Average-AP to evaluate shot spotting performance, 
    following the SoccerNet event spotting metric Average-mAP~\cite{Giancola2018}, computed over a single class in our setting. To account for small temporal misalignments, a prediction is considered correct if it falls within a predefined temporal tolerance around the ground-truth shooting frame. The final metric aggregates performance over multiple tolerance levels to provide a robust measure of spotting accuracy.
      
    Formally, assume the test set contains \(N\) samples
    \(\{\mathbf{X}_i\}_{i=1}^{N}\).
    We partition the samples into a positive set
    \(\mathcal{P}\), consisting of sequences that contain an event,
    and a negative set \(\mathcal{N}\), consisting of sequences with no events.
    Let \(N_+ = |\mathcal{P}|\).
    
    For each sample \(\mathbf{X}_i\), the model predicts an event timestamp
    \(\hat{t}_i\) with a confidence score \(\hat{s}_i\).
    For samples \(\mathbf{X}_i \in \mathcal{P}\), the ground-truth event occurs at frame
    \(t_i\).    
    Given a confidence threshold \(\theta\), a sample \(\mathbf{X}_i\) is predicted as positive at frame $\hat{t}_i$ if \(\hat{s}_i \ge \theta\).
    
    To allow minor temporal misalignment, we introduce a temporal tolerance
    \(\delta\).
    For samples in \(\mathcal{P}\), a prediction is considered temporally correct if
    $
    |\hat{t}_i - t_i| \le \delta.
    $
    
    Under tolerance \(\delta\), the number of true positives with a threshold \(\theta\)
    is
    \[
    \mathrm{TP}_\delta(\theta)
    =
    \sum_{\mathbf{X}_i \in \mathcal{P}}
    \mathbb{I}(\hat{s}_i \ge \theta)\,
    \mathbb{I}(|\hat{t}_i - t_i| \le \delta).
    \]
    
    The number of false positives is
    \[
    \mathrm{FP}_\delta(\theta)
    =
    \sum_{\mathbf{X}_i \in \mathcal{P}}
    \mathbb{I}(\hat{s}_i \ge \theta)\,
    \mathbb{I}(|\hat{t}_i - t_i| > \delta)
    +
    \sum_{\mathbf{X}_i \in \mathcal{N}}
    \mathbb{I}(\hat{s}_i \ge \theta).
    \]
              
    Precision and recall are then 
    % \[
    % P_\delta(\theta)
    % =
    % \frac{\mathrm{TP}_\delta(\theta)}
    % {\mathrm{TP}_\delta(\theta) + \mathrm{FP}_\delta(\theta)},
    % \\
    % R_\delta(\theta)
    % =
    % \frac{\mathrm{TP}_\delta(\theta)}{N_+}.
    % \]
    \[
    \begin{aligned}
    P_\delta(\theta)
    &=
    \frac{\mathrm{TP}_\delta(\theta)}
    {\mathrm{TP}_\delta(\theta)+\mathrm{FP}_\delta(\theta)},
    \\
    R_\delta(\theta)
    &=
    \frac{\mathrm{TP}_\delta(\theta)}{N_+}.
    \end{aligned}
    \]
    
    Performance under tolerance \(\delta\) is measured using Average Precision ($\delta$-AP),
    defined as the area under the precision--recall curve:
    \[
    \delta\text{-AP}
    =
    \int P_\delta(\theta)\, dR_\delta(\theta).
    \]
    
    In practice, the integral is computed using a step-wise approximation by
    sweeping the threshold \(\theta\).
    Let \(\pi\) denote the ordering of samples by decreasing confidence,
    \[
    \hat{s}_{\pi_1} \ge \hat{s}_{\pi_2} \ge \cdots \ge \hat{s}_{\pi_N}.
    \]
    Then \(\delta\)-AP can be written as
    % \[
    % \delta\text{-AP}
    % =
    % \sum_{i=1}^{N}
    % \bigl(
    % R_\delta(\hat{s}_{\pi_i})
    % -
    % R_\delta(\hat{s}_{\pi_{i-1}})
    % \bigr)\,
    % P_\delta(\hat{s}_{\pi_i}),
    % \quad
    % R_\delta(\hat{s}_{\pi_0}) = 0.
    % \]
    \[
    \begin{aligned}
    \delta\text{-AP}
    &=
    \sum_{i=1}^{N}
    \Bigl(
    R_\delta(\hat{s}_{\pi_i})
    -
    R_\delta(\hat{s}_{\pi_{i-1}})
    \Bigr)
    P_\delta(\hat{s}_{\pi_i}),
    \\
    &\qquad \text{where }
    R_\delta(\hat{s}_{\pi_0}) = 0.
    \end{aligned}
    \]
    
    Lowering the threshold from \(\hat{s}_{\pi_{i-1}}\) to \(\hat{s}_{\pi_i}\)
    includes exactly one additional prediction.
    Since only samples in \(\mathcal{P}\) may contribute to true positives,
    the corresponding increment in recall is
    \[
    R_\delta(\hat{s}_{\pi_i}) - R_\delta(\hat{s}_{\pi_{i-1}})
    =
    \frac{
    \mathbb{I}\bigl(\mathbf{X}_{\pi_i} \in \mathcal{P}\bigr)\,
    \mathbb{I}(|\hat{t}_{\pi_i} - t_{\pi_i}| \le \delta)
    }{
    N_+
    }.
    \]
    
    Substituting this expression yields the form of \(\delta\)-AP:
    \begin{equation*}
        \begin{aligned}
        \delta\text{-AP}
        &=
        \frac{1}{N_+}
        \sum_{i=1}^{N}
        P_\delta(\hat{s}_{\pi_i})\,
        \mathbb{I}\bigl(\mathbf{X}_{\pi_i} \in \mathcal{P}\bigr)\,
        \mathbb{I}(|\hat{t}_{\pi_i} - t_{\pi_i}| \le \delta) \\
        &=
        \frac{1}{N_+}
        \sum_{\mathbf{X}_{\pi_i} \in \mathcal{P}}
        P_\delta(\hat{s}_{\pi_i})\,
        \mathbb{I}(|\hat{t}_{\pi_i} - t_{\pi_i}| \le \delta).
        \end{aligned}
    \end{equation*}

    Finally, to summarize performance across different temporal tolerances, we
    report \textbf{Average-AP}, defined as the mean \(\delta\)-AP over a set of
    tolerances \(\Delta\):
    \[
    \mathrm{Average\mbox{-}AP}
    =
    \frac{1}{|\Delta|}
    \sum_{\delta \in \Delta}
    \delta\text{-AP}.
    \]
    In this work, we set $\Delta = \{0, 1, ..., 10\}$.

\begin{figure}[t]
    \centering
    \includegraphics[width=1.0\columnwidth]{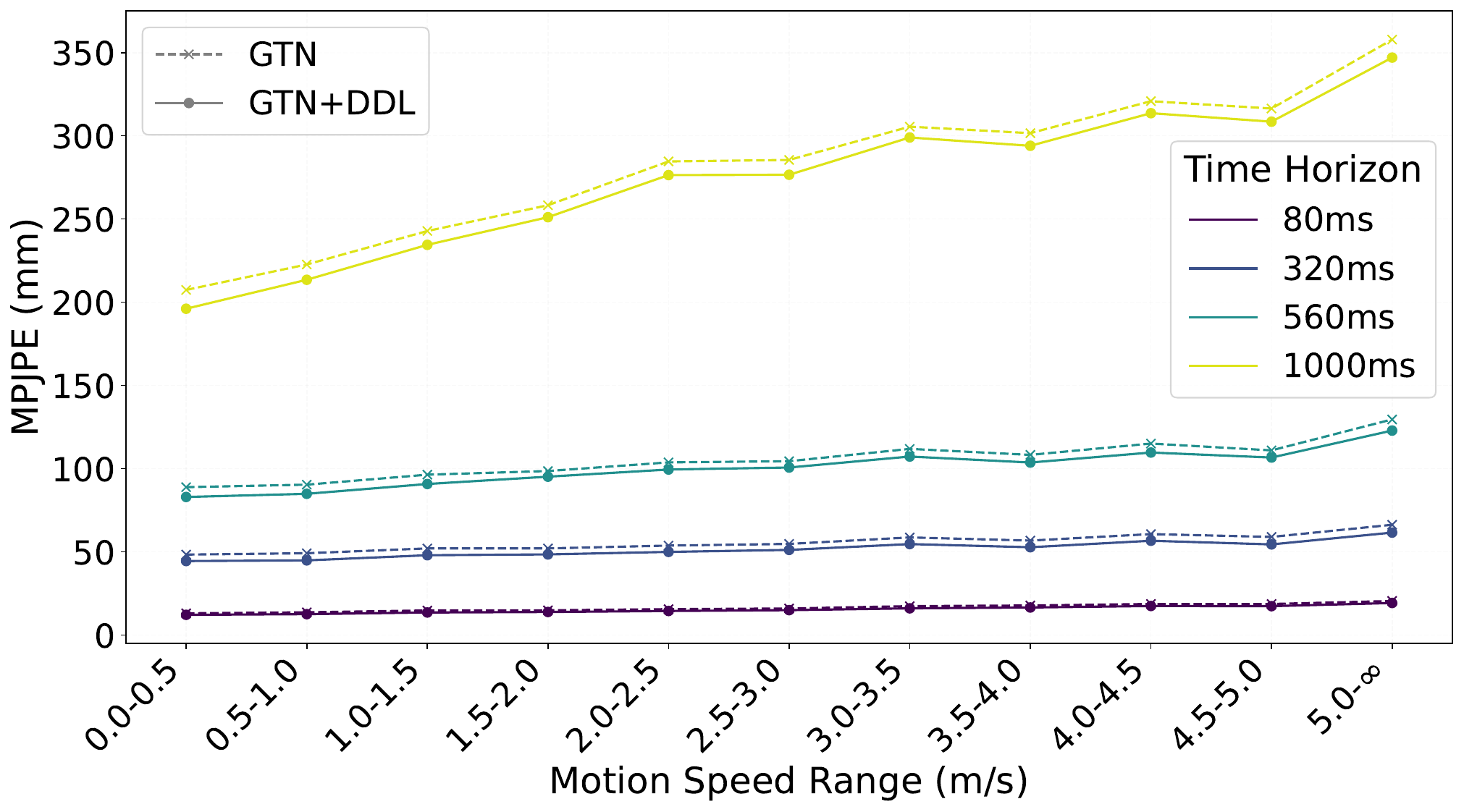}
    % \vspace{-5pt}
    \caption{\textit{Motion prediction performance across different motion speeds on ProSoccer.} We compare GTN with (solid) and without DDL (dashed).}
    \label{fig:range_eval_ProSoccer_gtn}
    % \vspace{-5pt}
\end{figure}

\begin{table*}[t]
\centering
% \vspace{-5pt}
\caption{\textit{Diverse motion prediction results with different temperatures.}}
\label{tab:diverse_metric_tau}
\resizebox{0.95\textwidth}{!}{%
\begin{tabular}{@{}l|lllll||lllll@{}}
\toprule
\multirow{2}{*}{Temperature $\tau$} & \multicolumn{5}{c||}{WorldPose}         & \multicolumn{5}{c}{ProSoccer}             \\
                                    & APD $\uparrow$   & ADE $\downarrow$    & FDE $\downarrow$   & MMADE $\downarrow$ & MMFDE $\downarrow$  & APD  $\uparrow$  & ADE $\downarrow$    & FDE $\downarrow$   & MMADE $\downarrow$ & MMFDE $\downarrow$ \\ \midrule
1                                   & 1.268 & 2.224 & 6.561 & 4.810 & 10.791 & 0.685 & 2.012 & 5.899 & 3.175 & 7.796 \\
2                                   & 1.873 & 2.171 & 6.344 & 4.751 & 10.552 & 1.318 & 1.929 & 5.574 & 3.088 & 7.452 \\
5                                   & 2.790 & 2.092 & 6.011 & 4.662 & 10.184 & 4.924 & 1.667 & 4.525 & 2.791 & 6.324 \\
10                                  & 3.615 & 2.021 & 5.718 & 4.582 & 9.852  & 5.147 & 1.687 & 4.560 & 2.796 & 6.320 \\ \bottomrule
\end{tabular}%
}
% \vspace{-5pt}
% \vspace{5pt}
\end{table*}

\section{Further Motion Prediction Results Across Different Speeds}
\label{appendix:range-eval-further}
        In addition to WorldPose, we also report motion prediction performance across different speed ranges on ProSoccer, as shown in Figure~\ref{fig:range_eval_ProSoccer}, comparing our GTN model trained with DDL against HisRep.
        Consistent with the WorldPose results in Figure \ref{fig:range_eval_worldpose}, our GTN+DDL model performs slightly worse than HisRep in the low-speed range but remains significantly more robust as motion speed increases. Consequently, it outperforms HisRep by an increasingly larger margin in higher-speed motion scenarios, demonstrating strong robustness to fast motion.

        Moreover, we compare our GTN trained with and without DDL across different motion speed ranges. Figure~\ref{fig:range_eval_worldpose_gtn} and~\ref{fig:range_eval_ProSoccer_gtn} show the results on WorldPose and ProSoccer, respectively. As shown, GTN and GTN+DDL exhibit similar trends as motion speed increases, with GTN+DDL outperforming the vanilla GTN in most cases by a notable margin, particularly when trained on the large ProSoccer dataset. These results suggest that much of the robustness to high-speed motion, compared with traditional models such as HisRep, comes from the frame-level learning framework with GTN, while DDL provides additional and consistent performance gains.

\section{Diverse Motion Prediction Results with \\Different Temperatures}
\label{appendix:diverse_results_with_taus}

DDL introduces stochasticity into motion prediction, with a temperature $\tau$ in Equation~\ref{eqn:dist} controlling the randomness of the sampling distribution. While $\tau$ is fixed during training, it can be adjusted at inference time to encourage more diverse predictions. In this section, we evaluate our model under different temperature settings with commonly used diverse motion prediction metrics.

% \noindent\textbf{Experiment Setting.}\quad
\paragraph{Experiment Setting.}
We evaluate our GTN+DDL models (last row in Table~\ref{tab:motion-prediction-merged}, both models trained on WorldPose and ProSoccer) under different temperatures during inference. 
% The model takes a 50-frame past motion (2 seconds) as input and autoregressively predicts the motion of 50 future frames (2 seconds). 
We use 50 input frames to autoregressively predict 50 future frames (2s $\rightarrow$ 2s), and generate $K = 50$ prediction samples per input.
We use subsets of the WorldPose and ProSoccer test sets for this experiment. For WorldPose, we include only the first two clips from the \texttt{BRA-KOR} match. For ProSoccer, we use approximately 2\% of the test set.

% \noindent\textbf{Metrics.}\quad
\paragraph{Metrics.}
We employ five commonly used metrics to evaluate both diversity (APD) and the accuracy of diverse motion predictions (ADE, FDE, MMADE, and MMFDE).
\begin{itemize}
    \item \textbf{APD}: Average Pairwise Distance (APD) measures sample diversity by computing the average $L_2$ distance between all pairs of motion samples. 
    \item \textbf{ADE}:  Average Displacement Error (ADE) calculates the average $L_2$ distance between the ground-truth future motion and the closest predicted motion sample over all time steps.
    \item \textbf{FDE}:  Final Displacement Error (FDE) computes the $L_2$ distance between the final ground-truth pose and the final pose of the closest predicted motion sample.
    \item \textbf{MMADE}: Multi-Modal ADE (MMADE) is the multimodal version of ADE. 
    Given multiple ground-truth future motions, MMADE is computed as the mean ADE over all ground-truth future motions.
    \item \textbf{MMFDE}: Multi-Modal FDE (MMFDE) is the multimodal version of FDE. 
    Given multiple ground-truth future motions, MMFDE is computed as the mean FDE over all ground-truth final poses.
\end{itemize}

Following prior works~\cite{Yuan2020DiverseProcesses,Xu2024LearningPrediction}, we obtain multimodal ground-truth motions by retrieving sequences with similar final observed poses, assuming that their ground-truth future motions are interchangeable as plausible continuations. Specifically, we retrieve sequences whose final observed poses are within an $L_2$ distance threshold of $0.5$, and use their future motions as alternative ground-truth futures.

% \noindent\textbf{Results.}\quad
\paragraph{Results.}
Table~\ref{tab:diverse_metric_tau} reports the diverse motion prediction results under different temperature settings on WorldPose and ProSoccer. Consistent with the qualitative results in Figure~\ref{fig:motion_prediction_visualization1}, increasing the temperature substantially improves diversity, as indicated by higher APD, while also improving the accuracy metrics. This trend is consistent across models trained on both WorldPose and ProSoccer. 
These results further suggest that our model captures motion multimodality through the discrete motion codebook. At $\tau=1$, the learned distribution appears sharp, likely due to the prediction-only training objective, as reflected by the limited APD. However, simply increasing the inference temperature better reveals the learned distribution, leading to more diverse predictions and improved performance across all metrics.

\section{Downstream Comparison with Pre-training Based Baselines}
\label{appendix:pretrain_baseline_downstream}

Although prior works (e.g., LTD~\cite{Mao2019LearningPrediction}, HisRep~\cite{Mao2020HistoryAttention}, and SiMLPe~\cite{Guo2023BackPrediction}) on motion prediction primarily focus on predictive accuracy and are not explicitly designed for representation learning, these methods nonetheless learn latent representations during training. In this section, we treat all our trained motion prediction baseline models in Table~\ref{tab:motion-prediction-merged} as pre-trained models and adapt them to downstream tasks in the same manner as our approach. Specifically, we extract representations from their latent space as learned representations, and use an identical fine-tuning protocol for both action recognition and shot spotting. We then evaluate and compare their performance with our models on these tasks. This allows the comparison with motion-prediction pre-training based baselines.
% , rather than restricting comparisons to non-pretraining methods.

\begin{table}[t]
% \vspace{-5pt}
\centering
\caption{\textit{Comparison of action recognition results with pre-training based baselines.}
``Random'': random initialization; 
``Pre-trained'': WorldPose (for WorldPoseAR) or ProSoccer (for SoccerAR) pre-trained model weights. 
Additional motion-prediction pre-training baselines are highlighted in red.
Our models are in blue.
}
\label{tab:action-recognition-extended}
\resizebox{\columnwidth}{!}{%
\begin{tabular}{@{}l|l|c|c@{}}
\toprule
\multirow{2}{*}{Model}
& \multirow{2}{*}{Initialization}
& WorldPoseAR
& SoccerAR \\
&
& Acc. (\%) $\uparrow$
& Acc. (\%) $\uparrow$ \\
\midrule

% Model & Initialization  & Acc. (\%) $\uparrow$ on WorldPoseAR & Acc. (\%) $\uparrow$ on SoccerAR  \\ \midrule
\multicolumn{4}{l}{\textit{From-scratch training methods}} \\ \midrule
ST-GCN~\cite{Yan2018}    & Random                   & \(94.66{\scriptstyle \pm 0.51}\) & \(55.81{\scriptstyle \pm 0.40}\) \\
CTR-GCN~\cite{Chen2021Channel-WiseRecognition}   & Random                   & \(92.42{\scriptstyle \pm 0.44}\) & \(55.31{\scriptstyle \pm 0.70}\) \\
BlockGCN~\cite{Zhou2024BlockGCN:Recognition}  & Random                   & \(78.96{\scriptstyle \pm 4.53}\) & \(57.87{\scriptstyle \pm 1.16}\) \\ 
\rowcolor{blue!8} GTN (ours) & Random                  & \(87.83{\scriptstyle \pm 0.36}\) & \(56.67{\scriptstyle \pm 0.96}\) \\ \midrule
\multicolumn{4}{l}{\textit{Pre-training based methods}} \\ \midrule
\rowcolor{red!8} LTD \cite{Mao2019LearningPrediction} & Pre-trained  (LTD)       & \(86.71{\scriptstyle \pm 1.16}\)  & \(49.94{\scriptstyle \pm 0.60}\) \\
\rowcolor{red!8} HisRep \cite{Mao2020HistoryAttention} & Pre-trained (HisRep)       & \(81.38{\scriptstyle \pm 5.06}\)  & \(51.20{\scriptstyle \pm 0.21}\) \\
\rowcolor{red!8} MSR-GCN \cite{Dang2021MSR-GCN:Prediction} & Pre-trained (MSR-GCN)       & \(75.75{\scriptstyle \pm 1.65}\)  & \(40.63{\scriptstyle \pm 0.45}\) \\
\rowcolor{red!8} PGBIG \cite{Ma2022ProgressivelyPrediction} & Pre-trained (PGBIG)       & \(70.08{\scriptstyle \pm 2.02}\)  & \(35.87{\scriptstyle \pm 0.33}\) \\
\rowcolor{red!8} SiMLPe \cite{Guo2023BackPrediction} & Pre-trained (SiMLPe)       & \(70.33{\scriptstyle \pm 2.25}\)  & \(37.31{\scriptstyle \pm 1.04}\) \\
\rowcolor{red!8} GCNext \cite{Wang2024GCNext:Prediction} & Pre-trained (GCNext)       & \(77.83{\scriptstyle \pm 2.75}\)  & \(42.73{\scriptstyle \pm 0.96}\) \\ \midrule
MAMP~\cite{Mao2023MaskedLearners} & Pre-trained (MAMP)      & \(93.17{\scriptstyle \pm 0.50}\) & \(59.12{\scriptstyle \pm 0.18}\)  \\
\rowcolor{blue!8} GTN (ours) & Pre-trained (GTN)       & \(95.41{\scriptstyle \pm 0.75}\) & \(62.70{\scriptstyle \pm 1.01}\) \\
\rowcolor{blue!8} GTN (ours) & Pre-trained (GTN+DDL) & \(\textbf{96.21}{\scriptstyle \pm 0.29}\) & \(\textbf{64.15}{\scriptstyle \pm 0.47}\) \\ \bottomrule
\end{tabular}%
}
\vspace{-10pt}
\end{table}

    \subsection{Action Recognition}
    \label{appendix:extended-action-recognition}

    Table~\ref{tab:action-recognition-extended} presents action recognition results in comparison with motion-prediction pre-training based methods. The methods are grouped into from-scratch and pre-training based categories. All motion-prediction pre-training baselines (highlighted in red) perform substantially worse than our models and remain far behind non-pretraining baselines such as ST-GCN. This suggests that, despite sharing the same motion prediction objective, differences in model design and learning framework lead to representations with varying degrees of transferability, with our method being significantly more effective for downstream tasks like action recognition.

    \subsection{Shot Spotting}
    \label{appendix:extended-shot-spotting}

    Table~\ref{tab:shot-spotting-extended} shows shot spotting results in comparison with motion-prediction pre-training based methods. Methods are grouped into from-scratch and pre-training based categories. As can be seen from the table, motion-prediction pre-training based baselines (highlighted in red) yield substantially inferior results, failing to match the performance of from-scratch training methods and our models. A likely reason is that these methods typically rely on discrete cosine transform (DCT) and are trained at the sequence level, which leads to oversmoothed frame-level representations. In contrast, our model benefits from explicit fine-grained frame-level modeling.
    This further suggests that motion prediction alone is insufficient for transferability. Our model design and learning framework are essential for effective representation learning, with the frame-level objective further benefiting fine-grained tasks such as shot spotting.
    
\begin{table}[t]
% \vspace{-10pt}
\centering
\caption{\textit{Comparison of shot spotting results with pre-training based baselines.} 
``Random'': random initialization; 
``Pre-trained'': ProSoccer pre-trained model weights. 
Additional motion-prediction pre-training baselines are highlighted in red.
Our models are in blue.
}
\label{tab:shot-spotting-extended}
\resizebox{0.88\columnwidth}{!}{%
\begin{tabular}{@{}l|l|l@{}}
\toprule
Model    & Initialization              & \multicolumn{1}{c}{Average-AP $\uparrow$} \\ \midrule
\multicolumn{3}{l}{\textit{From-scratch training methods}} \\ \midrule
ST-GCN \cite{Yan2018}  
& Random                           & \(0.8776{\scriptstyle \pm 0.0070}\)                         \\
CTR-GCN \cite{Chen2021Channel-WiseRecognition}  
& Random                           & \(0.8835{\scriptstyle \pm 0.0010}\)                        \\
BlockGCN \cite{Zhou2024BlockGCN:Recognition} 
& Random                           & \(0.8540{\scriptstyle \pm 0.0324}\)                         \\ 
\rowcolor{blue!8}GTN (ours)     & Random                           & \(0.8957{\scriptstyle \pm 0.0022}\)                           \\ \midrule
\multicolumn{3}{l}{\textit{Pre-training based methods}} \\ \midrule
\rowcolor{red!8} LTD \cite{Mao2019LearningPrediction} & Pre-trained (LTD)       & \(0.7166{\scriptstyle \pm 0.0212}\)  \\
\rowcolor{red!8} HisRep \cite{Mao2020HistoryAttention} & Pre-trained (HisRep)       & \(0.7265{\scriptstyle \pm 0.0046}\)  \\
\rowcolor{red!8} MSR-GCN \cite{Dang2021MSR-GCN:Prediction} & Pre-trained (MSR-GCN)       & \(0.4351{\scriptstyle \pm 0.0208}\)  \\
\rowcolor{red!8} PGBIG \cite{Ma2022ProgressivelyPrediction} & Pre-trained (PGBIG)       & \(0.4236{\scriptstyle \pm 0.0021}\)  \\
\rowcolor{red!8} SiMLPe \cite{Guo2023BackPrediction} & Pre-trained (SiMLPe)       & \(0.6985{\scriptstyle \pm 0.0098}\)  \\
\rowcolor{red!8} GCNext \cite{Wang2024GCNext:Prediction} & Pre-trained (GCNext)       & \(0.6821{\scriptstyle \pm 0.0057}\) \\ \midrule
MAMP~\cite{Mao2023MaskedLearners} & Pre-trained (MAMP)   & \(0.8237{\scriptstyle \pm 0.0031}\) \\
\rowcolor{blue!8}GTN (ours)     & Pre-trained (GTN)            & \(0.9242{\scriptstyle \pm 0.0035}\)                          \\
\rowcolor{blue!8}GTN (ours)     & Pre-trained (GTN+DDL) & \(\textbf{0.9274}{\scriptstyle \pm 0.0030}\)                \\ \bottomrule
\end{tabular}%
}
% \vspace{-10pt}
\vspace{-5pt}
\end{table}

\begin{figure}[t]
    \centering
    \includegraphics[width=0.95\linewidth]{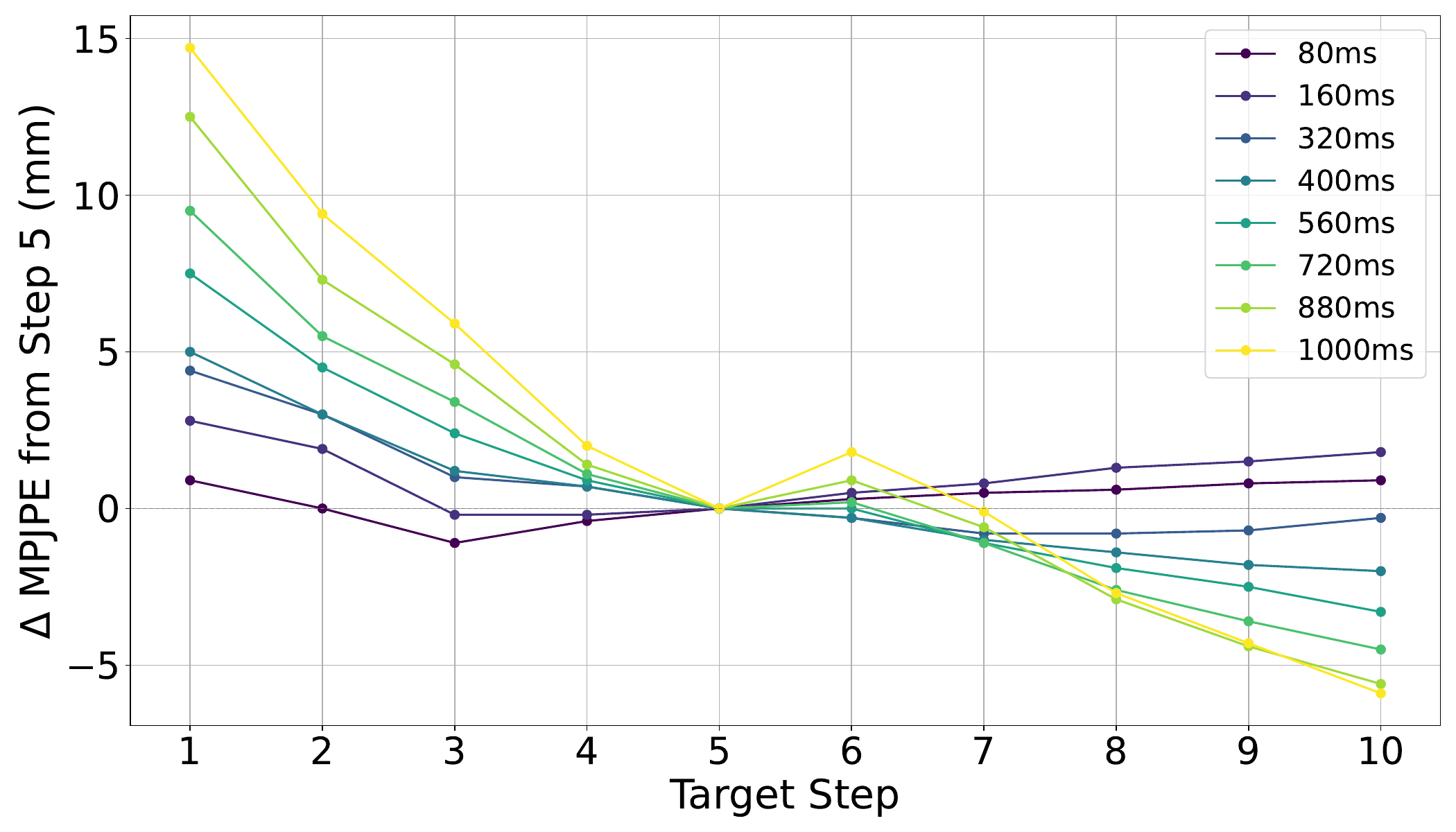}
    \caption{\textit{Impact of the target step on ProSoccer.}  The plot shows MPJPE changes relative to the default setting (the 5th step), where positive $\Delta$MPJPE indicates worse performance than the default setting and negative values indicate improvement.}
    \label{fig:step_ablation_ProSoccer}
\vspace{-10pt}
\end{figure}

\section{Further Ablation Studies}

    \subsection{Impact of the Target Step on ProSoccer}
    \label{appendix:ablation-step}

        Figure~\ref{fig:step_ablation_ProSoccer} illustrates the effect of selecting different future motion steps as the prediction target for distribution learning on ProSoccer. Consistent with the WorldPose results in Figure~\ref{fig:step_ablation_worldpose}, choosing an earlier step (e.g., the 3rd step) improves short-term prediction but degrades long-term performance, while selecting a later step (e.g., the 8th-10th steps) enhances long-term prediction at the expense of short-term accuracy. An exception is the 1st step, which yields suboptimal results, likely due to its strong dependence on motion history and the limited multimodality at that time horizon. Overall, consistent with the findings on WorldPose, using the 5th step as the prediction target provides a balanced trade-off between short- and long-term performance.

\begin{table*}[t]
\centering
\caption{\textit{Ablation study on codebook design.}}
\vspace{-5pt}
\label{tab:ablation-codebook-merged}
\resizebox{\linewidth}{!}{%
\begin{tabular}{@{}l|cccccccc||cccccccc@{}}
\toprule
 & \multicolumn{8}{c||}{MPJPE (mm) $\downarrow$ on WorldPose} & \multicolumn{8}{c}{MPJPE (mm) $\downarrow$ on ProSoccer} \\
\cmidrule(r){2-9} \cmidrule(l){10-17}
 & 80ms & 160ms & 320ms & 400ms & 560ms & 720ms & 880ms & 1000ms
 & 80ms & 160ms & 320ms & 400ms & 560ms & 720ms & 880ms & 1000ms \\
\midrule
\textbf{Full model}
 & \textbf{15.6} & \textbf{32.5} & \textbf{61.0} & \textbf{76.2} & \textbf{115.9} & 173.4 & 256.7 & 338.6
 & \textbf{14.1} & \textbf{24.9} & \textbf{49.6} & \textbf{63.0} & \textbf{95.2} & \textbf{137.1} & \textbf{194.1} & \textbf{248.0} \\
\midrule
w/o nonlinear projection
 & 15.7 & 32.7 & 61.2 & 76.3 & 116.3 & 174.1 & 258.2 & 341.0
 & 14.2 & 25.1 & 50.0 & 63.6 & 96.2 & 138.5 & 196.3 & 251.2 \\
w/o weight tying
 & \textbf{15.6} & \textbf{32.5} & 61.2 & 76.4 & 116.5 & 175.0 & 259.8 & 342.8
 & \textbf{14.1} & 25.1 & 50.0 & 63.4 & 96.0 & 138.6 & 196.4 & 251.2 \\
w/o sampling during training
 & \textbf{15.6} & 32.6 & 61.2 & 76.3 & 116.3 & \textbf{173.2} & \textbf{256.1} & \textbf{336.9}
 & \textbf{14.1} & 25.0 & 49.8 & 63.2 & 95.5 & 137.6 & 194.6 & 248.8 \\
\bottomrule
\end{tabular}
}
\vspace{-5pt}
\end{table*}

\begin{table*}[t]
% \vspace{-10pt}
\centering
\caption{\textit{Impact of codebook size.}}
\vspace{-5pt}
\label{tab:hyperparam-vocabsize-merged}
\resizebox{\linewidth}{!}{%
\begin{tabular}{@{}l|cccccccc||cccccccc@{}}
\toprule
 & \multicolumn{8}{c||}{MPJPE (mm) $\downarrow$ on WorldPose} & \multicolumn{8}{c}{MPJPE (mm) $\downarrow$ on ProSoccer} \\
\cmidrule(r){2-9} \cmidrule(l){10-17}
\# Codes
 & 80ms & 160ms & 320ms & 400ms & 560ms & 720ms & 880ms & 1000ms
 & 80ms & 160ms & 320ms & 400ms & 560ms & 720ms & 880ms & 1000ms \\
\midrule
16
 & 15.8 & 33.1 & 61.6 & 76.6 & 116.6 & 174.7 & 258.4 & 340.8
 & 14.6 & 26.5 & 51.8 & 65.2 & 97.8 & 140.3 & 198.2 & 253.4 \\
32
 & 15.9 & 33.2 & 62.0 & 77.0 & 117.2 & 175.9 & 260.9 & 344.8
 & 14.5 & 26.2 & 51.4 & 64.7 & 97.2 & 139.5 & 197.4 & 252.7 \\
64
 & 15.7 & 32.7 & 61.3 & 76.4 & 116.4 & 174.3 & 258.0 & 340.8
 & 14.4 & 25.8 & 50.7 & 64.0 & 96.5 & 138.7 & 196.2 & 251.1 \\
128
 & 15.7 & 32.9 & 62.1 & 77.6 & 118.0 & 176.6 & 261.3 & 344.4
 & 14.3 & 25.5 & 50.4 & 63.8 & 96.5 & 139.1 & 197.1 & 252.5 \\
256
 & 15.6 & 32.5 & 61.3 & 76.5 & 116.4 & 173.7 & 256.4 & 337.0
 & 14.2 & 25.2 & 50.0 & 63.4 & 96.1 & 138.6 & 196.5 & 251.4 \\
512
 & 15.6 & 32.5 & 61.0 & \textbf{76.2} & \textbf{115.9} & \textbf{173.4} & \textbf{256.7} & \textbf{338.6}
 & 14.1 & 24.9 & 49.6 & 63.0 & 95.2 & 137.1 & 194.1 & 248.0 \\
1024
 & 15.6 & 32.4 & 61.1 & 76.5 & 116.7 & 175.3 & 259.4 & 342.7
 & 14.0 & 24.7 & \textbf{49.4} & 63.0 & 96.0 & 138.9 & 196.8 & 251.9 \\
2048
 & \textbf{15.5} & 32.3 & 61.2 & 76.5 & 116.8 & 175.7 & 259.4 & 341.8
 & 13.9 & 24.6 & 49.7 & 63.0 & \textbf{95.1} & \textbf{136.8} & \textbf{193.8} & \textbf{247.7} \\
4096
 & 15.6 & 32.3 & \textbf{60.9} & \textbf{76.2} & 116.0 & 174.7 & 257.5 & 339.4
 & \textbf{13.8} & \textbf{24.3} & 49.5 & 63.0 & 95.4 & 137.6 & 195.1 & 249.8 \\
8192
 & 15.6 & \textbf{32.1} & 61.0 & 76.4 & 116.9 & 175.7 & 259.7 & 343.3
 & \textbf{13.8} & \textbf{24.3} & 49.5 & \textbf{62.9} & 95.7 & 138.4 & 196.2 & 251.1 \\
\bottomrule
\end{tabular}
}
\vspace{-10pt}
\end{table*}

    \subsection{Ablation Study on Codebook Design}
    \label{appendix:ablation-codebook}
    
        We perform ablation studies on our discrete codebook design by removing or replacing specific components. Specifically, we evaluate 3 model variants:
        \begin{itemize}
            \item \textbf{w/o nonlinear projection}: The MLP layer before code prediction is replaced by a linear layer, excluding nonlinearity.
            \item \textbf{w/o weight tying}: The weights of the code embedding and the prediction layer are untied.
            \item \textbf{w/o sampling during training}: Sampling during training is disabled. Instead, we directly select the code with the highest probability to guide the final motion prediction during training.
        \end{itemize}
        In Table~\ref{tab:ablation-codebook-merged}, we present results on both WorldPose and ProSoccer. 
        Removing the nonlinearity in the code prediction layer (row 2) consistently degrades performance. 
        This is likely because the same node representation is linearly mapped to both code classification and motion prediction, forcing it to serve two competing objectives and limiting task-specific specialization.
        In the third row of the table, we untie the weights of the code embedding and the prediction layer. Although this introduces slightly more model parameters, the performance degrades, suggesting that sharing weights between the code embedding and the prediction layer leads to more efficient training.
        Disabling sampling during training, as shown in the last row of the table, results in only a marginal performance drop, indicating that sampling is not a critical factor for overall performance.

    \subsection{Impact of Codebook Size $K$} 
    \label{appendix:ablation-hyperparameter}

        % \paragraph{Impact of Codebook Size $K$} 
        We also investigate the impact of codebook size $K$. In Table \ref{tab:hyperparam-vocabsize-merged}, we report results using codebook sizes ranging from 16 to 8192 on both WorldPose and ProSoccer. As shown in the table, increasing the codebook size from 16 to 512 consistently improves performance across all time horizons on both datasets. However, beyond 512, further increases provide no observable benefit on WorldPose, and only minor improvements for short-term prediction on ProSoccer, while long-term performance shows no notable difference. Given the increased computational and memory costs with larger codebooks, a codebook size of 512 represents a favorable trade-off between efficiency and accuracy.

    \subsection{Impact of Frame-level Learning on \\Frame-level Applications}
    \label{appendix:ablation-frame-level-application}
    
        % To assess the importance of frame-level pre-training for frame-level downstream tasks, we pre-train a GTN+DDL variant using sequence-level supervision and fine-tune it for shot spotting under the same protocol as the full model.
        To assess the importance of frame-level pre-training for frame-level downstream tasks, we reuse the ProSoccer GTN+DDL model trained with sequence-level supervision in the ``w/o frame-level'' ablation of Table~\ref{tab:ablation-joint-comparison} and fine-tune it for shot spotting under the same protocol as the full model.
        As shown in Table~\ref{tab:ablation-frame-level-application}, replacing frame-level supervision with sequence-level supervision during pre-training decreases Average-AP for shot spotting by $0.0132$. This gap demonstrates that frame-level pre-training yields more temporally fine-grained representations that transfer more effectively to frame-level tasks such as shot spotting.

    \begin{table}[t]
        \centering
        \caption{
        \textit{Impact of frame-level pre-training on shot spotting.}
        % Both models are pre-trained on ProSoccer and fine-tuned under the same protocol.
        % Higher is better.
        }
        \label{tab:ablation-frame-level-application}
        \resizebox{\columnwidth}{!}{%
        \begin{tabular}{l|l|l@{}}
            \toprule
            Model & Initialization & Average-AP $\uparrow$ \\
            \midrule
            GTN (ours) & Pre-trained (sequence-level GTN+DDL)
                & $0.9142{\scriptstyle \pm 0.0033}$ \\
            GTN (ours) & Pre-trained (frame-level GTN+DDL)
                & $\mathbf{0.9274}{\scriptstyle \pm 0.0030}$ \\
            \bottomrule
        \end{tabular}%
        }
    \end{table}

% \section{Broader Impacts}
% \label{appendix:broader_impacts}

% The goal of this work is to advance human motion understanding with machine learning in the context of soccer. Its potential positive impacts include improved sports analytics, enabling players, coaches, and clubs to evaluate performance from a data-driven perspective. It may also support the automation of game data production, reducing reliance on manual annotation and lowering barriers for smaller leagues and clubs with limited resources.
% More broadly, the proposed future distribution learning paradigm is not limited to soccer player motion and may benefit a wide range of applications involving uncertain and multimodal human motion.

% Potential negative impacts include the misuse of predictive models, which are inherently imperfect. Inaccurate or biased predictions may lead to misleading analyses or flawed decision-making. In addition, relying primarily on data from professional men’s soccer may introduce representation bias, limiting the generalizability of the learned representations to women’s soccer or amateur play. These risks highlight the importance of careful validation and responsible use in downstream applications.